\documentclass[pdflatex,sn-mathphys-num]{sn-jnl}

\usepackage{xcolor}
\usepackage{graphicx}%
\usepackage{multirow}%
\usepackage{amsmath,amssymb,amsfonts}%
\usepackage{amsthm}%
\usepackage{mathrsfs}%
\usepackage[title]{appendix}%
\usepackage{xcolor}%
\usepackage{textcomp}%
\usepackage{manyfoot}%
\usepackage{booktabs}%
\usepackage{algorithm}%
\usepackage{algorithmicx}%
\usepackage{algpseudocode}%
\usepackage{listings}%
\usepackage{subcaption} 
\usepackage{bm}
\usepackage[capitalise]{cleveref}
\usepackage{lineno}
\usepackage[font=small,labelfont=bf,figurename=Fig.,labelsep=space]{caption}

\begin{document}

\title[Automatic Multi-View X-Ray/CT Registration Using Bone Substructure Contours]{Automatic Multi-view X-ray/CT Registration using Bone Substructure Contours}

\author*[1,2]{\fnm{Roman} \sur{Flepp}}\email{roman.flepp@kispi.uzh.ch}
\author[2]{\fnm{Leon} \sur{Nissen}}
\author[2]{\fnm{Bastian} \sur{Sigrist}}
\author[3]{\fnm{Arend} \sur{Nieuwland}}
\author[2]{\fnm{Nicola} \sur{Cavalcanti}}
\author[2]{\fnm{Philipp} \sur{Fürnstahl}}
\author[1,3]{\fnm{Thomas} \sur{Dreher}}
\author[2]{\fnm{Lilian} \sur{Calvet}}

\affil[1]{\orgname{University Children's Hospital Zürich}, \orgaddress{\country{Switzerland}}}
\affil[2]{\orgdiv{Research in Orthopedic Computer Science}, \orgname{University Hospital Balgrist, University of Zurich}, \orgaddress{\country{Switzerland}}}
\affil[3]{\orgdiv{Department of Orthopedic Surgery}, \orgname{University Hospital Balgrist, University of Zurich}, \orgaddress{ \country{Switzerland}}}

\abstract{

\textbf{Purpose:} Accurate intraoperative X-ray/CT registration is essential for surgical navigation in orthopedic procedures. However, existing methods struggle with consistently achieving sub-millimeter accuracy, robustness under broad initial pose estimates or need manual key-point annotations. This work aims to address these challenges by proposing a novel multi-view X-ray/CT registration method for intraoperative bone registration.

\noindent\textbf{Methods:} The proposed registration method consists of a multi-view, contour-based iterative closest point (ICP) optimization. Unlike previous methods, which attempt to match bone contours across the entire silhouette in both imaging modalities, we focus on matching specific subcategories of contours corresponding to bone substructures. This leads to reduced ambiguity in the ICP matches, resulting in a more robust and accurate registration solution.
This approach requires only two X-ray images and operates fully automatically. Additionally, we contribute a dataset of 5 cadaveric specimens, including real X-ray images, X-ray image poses and the corresponding CT scans.

\noindent\textbf{Results:} The proposed registration method is evaluated on real X-ray images using mean reprojection error (mRPD). The method consistently achieves sub-millimeter accuracy with a mRPD 0.67mm compared to 5.35mm by a commercial solution requiring manual intervention. Furthermore, the method offers improved practical applicability, being fully automatic.

\noindent\textbf{Conclusion:} Our method offers a practical, accurate, and efficient solution for multi-view X-ray/CT registration in orthopedic surgeries, which can be easily combined with tracking systems. By improving registration accuracy and minimizing manual intervention, it enhances intraoperative navigation, contributing to more accurate and effective surgical outcomes in computer-assisted surgery (CAS). The source code and the dataset are publicly available at: \url{https://github.com/rflepp/MultiviewXrayCT-Registration}.}


\maketitle

\section{Introduction}

Medical imaging technologies, such as X-ray and Computed Tomography (CT) scans, are critical for orthopedic diagnostics, surgical planning, and intraoperative guidance \cite{surgery_computer_vision}.
Aligning intraoperative X-ray images with preoperative CT scans provides surgeons with enhanced visualization and serves as a fundamental component of surgical navigation systems \cite{furman_AR}. This alignment enables various functionalities, such as the localization of surgical tools relative to the patient's anatomy, thereby improving surgical precision and patient outcomes \cite{Ahmad2014}. However, existing intraoperative registration techniques face significant limitations. 
For instance, intensity-based methods are highly sensitive to initial conditions, where even a few millimeters of error in the initial positioning can cause the algorithm to converge to a local minimum \cite{gopalakrishnan2024intraoperative2d3dimageregistration}. Other methods suffer from cumbersome workflows or manual keypoint extraction. \cite{ liao2020multiview2d3drigidregistration, ImFusion}.

X-ray/CT registration methods generally fall into two categories: intensity-based and landmark-based. Intensity-based methods \cite{Grupp_2020, ImFusion} optimize similarity between digitally reconstructed radiographs (DRR) and acquired X-ray images. Traditional optimization using non-differentiable DRR renderers \cite{Grupp_2020} are often computationally slow. Recently proposed differentiable methods like DiffPose \cite{gopalakrishnan2024intraoperative2d3dimageregistration} alleviate this issue. Nonetheless, these methods remain sensitive to initial conditions \cite{Grupp_2020}. Landmark-based methods, use identifiable anatomical points, contours, or external markers to establish correspondences between the two imaging modalities. However, these methods can become unreliable due to ambiguity in landmark extraction or when too few landmarks are available \cite{Grupp_2020_2}.
Contour-based methods, without strong semantic cues, poorly constrain the registration solution. Similar contours can arise from different anatomy poses, leading to ambiguity \cite{zhang_contour}.
 Recently, deep learning has enabled data-driven registration techniques in both categories \cite{gopalakrishnan2024intraoperative2d3dimageregistration}, learning the correct mappings and providing good initializations. However, these methods often require large amounts of data and computing power and struggle to generalize across different patients or imaging conditions.

Intraoperative X-ray/CT registration can be multi-view or single-view. Multi-view registration generally offers higher accuracy \cite{Grupp_2020_2, multi_view_registration_2011, liao2020multiview2d3drigidregistration}. However, it involves an additional calibration step, typically performed using in-image markers \cite{multi_view_registration_2011}. Calibration accuracy is crucial, as any errors will propagate through the entire registration system.

We present a novel, fully automated multi-view X-ray/CT registration method designed for intraoperative use, with a focus on both accuracy and ease of use. Our approach enhances classical contour-based methods by leveraging anatomical substructures, which improves the robustness and makes the optimization problem better posed, eliminating the need for manual keypoint extraction. Focused on the femur bone, the technique is evaluated on real data. Additionally, we introduce a publicly available dataset consisting of 5 cadaveric femur CT scans with corresponding X-ray images, including contour labels and precise pose data, supporting further research in intraoperative multi-view X-ray/CT registration.

The rest of this work is organized as follows. Section~\ref{sec:sota} discusses state-of-the-art X-ray/CT registration methods. The proposed registration method is presented in Section~\ref{sec:method}. Section~\ref{sec:xp} extensively evaluates the proposed method, before our conclusions are drawn in Section~\ref{sec:conclusion}.

\section{Related Work}
\label{sec:sota}

\noindent\textbf{Intensity-Based Registration} Intensity-based registration methods align digitally reconstructed radiographs (DRR), which are computed using the preoperative CT, with the measured X-ray by optimizing a similarity measure between image intensities, such as Mutual Information (MI) or Normalized Cross-Correlation (NCC) in single-view registration \cite{Grupp_2020} and multi-view registration \cite{ImFusion, multi_view_registration_2011}.
These methods can achieve high accuracy and their performance is agnostic to the anatomy shape, making them broadly applicable. However, they are often computationally intensive and sensitive to the initial alignment due to the non-convex nature of the optimization landscape and thus make use of landmark detection for initialization \cite{liao2020multiview2d3drigidregistration, gopalakrishnan2024intraoperative2d3dimageregistration}. Some methods aim to optimize computational efficiency while maintaining acceptable accuracy \cite{precision_timing_2}. Additionally, modality differences between X-ray and CT images can lead to varying intensity distributions, making it challenging for these methods to find a global optimum without good initialization.

\noindent\textbf{Landmark-Based Registration} Landmark-based methods utilize identifiable anatomical points, contours or external markers to establish correspondence between images in single-view \cite{Grupp_2020_2} and multi-view settings \cite{liao2020multiview2d3drigidregistration, zhang_contour}. The Perspective-n-Point (PnP) algorithm is then applied for X-ray image pose estimation from the extracted landmarks. These methods rely heavily on the accurate detection of landmarks, which can be hard to extract accurately and can be ambiguous, or insufficient in number, depending on the extraction strategy used.

\noindent\textbf{Learning-Based Methods} Deep learning approaches have been employed to predict transformations directly from a single X-ray \cite{miao2016realtime2d3dregistrationcnn, gopalakrishnan2024intraoperative2d3dimageregistration} and a set of posed X-ray images \cite{liao2020multiview2d3drigidregistration}. Deep learning models can learn to estimate the registration parameters \cite{miao2016realtime2d3dregistrationcnn, gopalakrishnan2024intraoperative2d3dimageregistration} or point correspondences \cite{liao2020multiview2d3drigidregistration} but require large amounts of annotated data and may not generalize well to different patients or imaging conditions due to the domain gap of synthetic and real X-ray images. Nonetheless, the popularity has increased in learning-based methods over the last years.

\noindent \textbf{Contour-Based 2D-3D Registration} Several contour-based approaches have been proposed for 2D/3D registration \cite{contour_based, zhang_contour}. These methods aim to align 3D models with 2D X-ray images or laparoscopic images by minimizing the reprojection error between the projected contours from 3D models (typically derived from CT or MRI) and the contours extracted from the 2D images. Techniques such as Iterative Closest Point (ICP) are often used to refine the alignment \cite{zhang_contour}. However, the problem is often ill-posed, and its solution suffers from ambiguities in matching 2D contours to the 3D model, as well as sensitivity to initial alignment. In contrast to these methods that use object-level contours, typically the bone silhouette, we propose leveraging recent advances in deep learning-based semantic contour extraction to extract component-level contours, specifically the contours of its anatomical substructures. Matching contours within the same class of substructure helps reduce ambiguities.
This results in a registration solution that is initialization-free, offering greater accuracy and robustness.

\section{Methodology}
\label{sec:method}

\begin{figure}[h]
    \centering
    \includegraphics[width=0.99\linewidth]{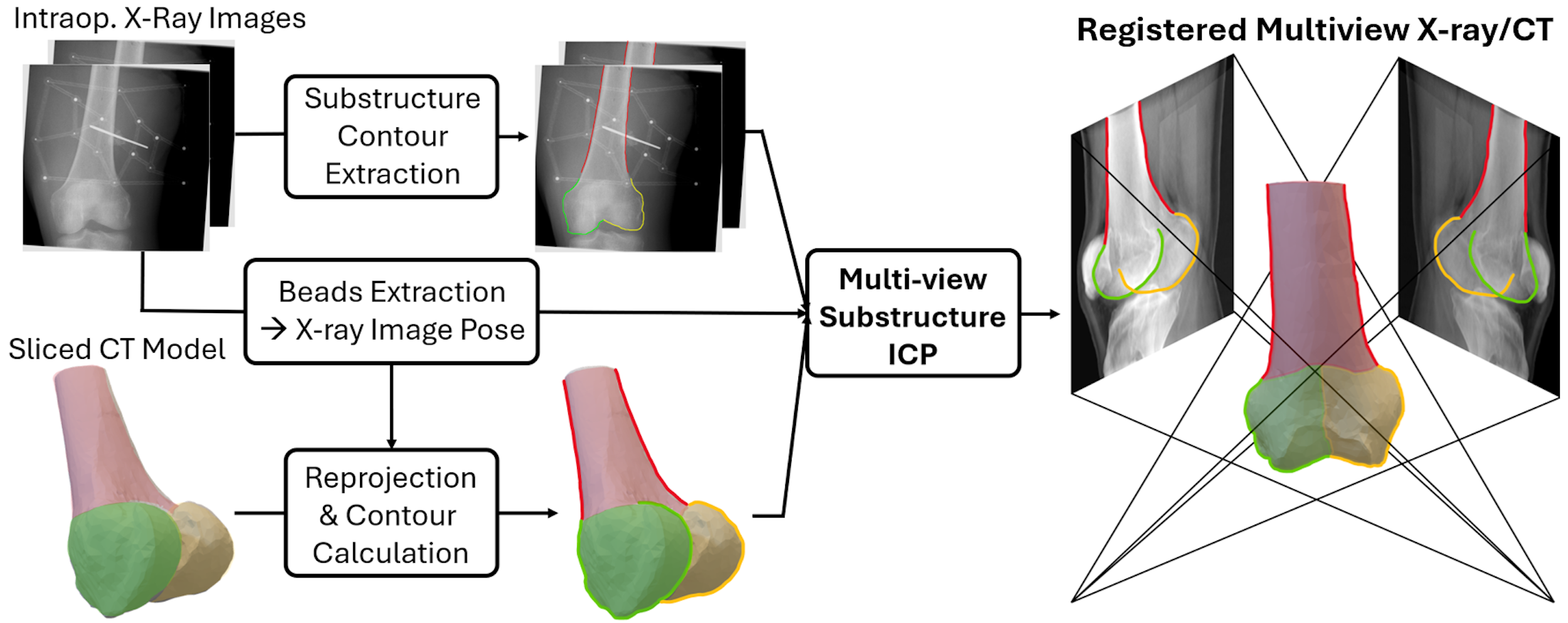}
    \caption{Overview of the proposed registration pipeline. X-ray image poses are computed using a fiducial marker. Contours of the diaphysis (red), medial and lateral condyles (green and yellow resp.) are automatically extracted in the images. The registration is performed using a contour-based ICP with matches across the same classes of substructures
 }
    \label{fig:pipeline}
\end{figure}

The proposed X-ray/CT calibration method is as follows.
The input X-ray images are calibrated using a custom-made fiducial marker, which is attached to a 3 mm-diameter K-wire fixed to the femur.
Occluding contours of the femur substructures, specifically the diaphysis and condyles, are extracted using a U-Net-based model trained on synthetic specimen-specific data.
The contour segmentation process is detailed in Section \ref{sec:model}.
Finally, the X-ray/CT registration is solved by applying a contour-based ICP optimization to the predicted contours, as described in Section \ref{sec:ICP}.
An overview of our registration method is shown in Figure~\ref{fig:pipeline}.

\begin{figure}[h]
    \centering
    \includegraphics[width=0.9\linewidth]{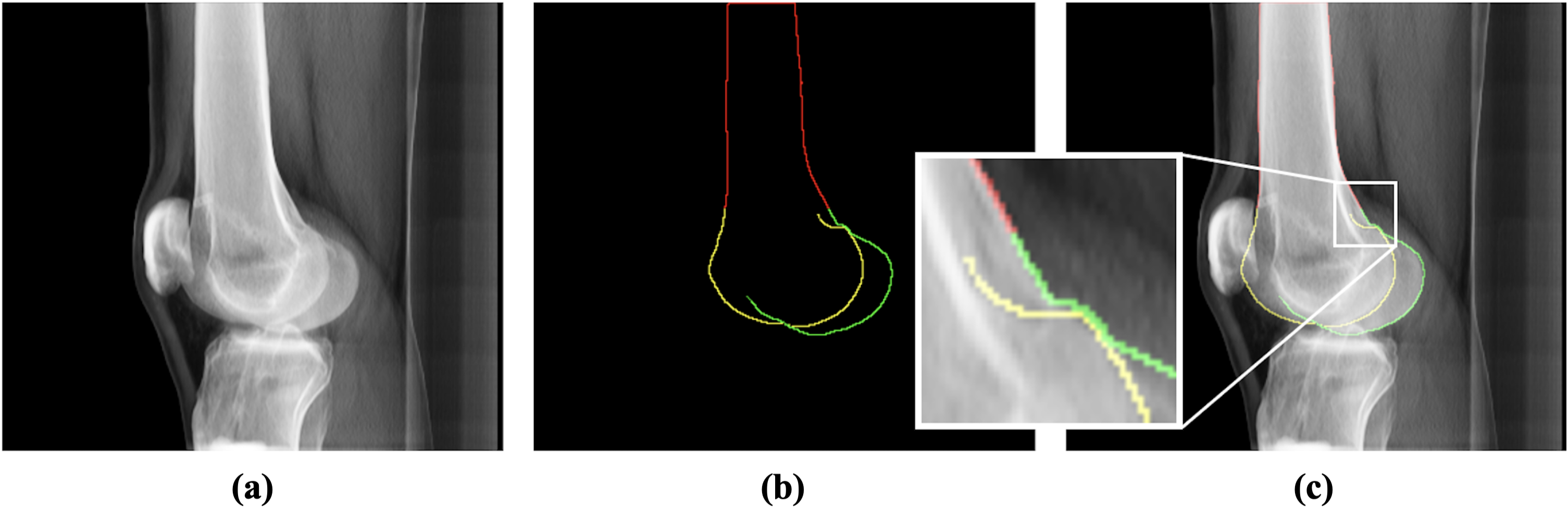}
    \caption{Example of a training sample. \textbf{(a)}: DRR. \textbf{(b)}: contour labels of the diaphysis (red), medial and lateral condyles (green and yellow resp.). \textbf{(c)}: overlay of both}
    \label{fig:training_data_synth}
\end{figure}

\subsection{Semantic Contour Segmentation}
\label{sec:model}

The key idea behind our contour-based ICP registration method is to achieve accurate 2D/3D matches using component-level contours instead of object-level ones. This raises the question of how to define component classes based on the anatomy of the femur.
Although X-rays are transparent, for the sake of clarity, we assume the femur to be an opaque object for now and focus specifically on its occluding contours, which appear in the X-ray as sudden changes in tissue density. We recall that for an occluding contour to form, the shape must be locally convex in the viewing direction. In other words, at the point where the occluding contour is observed, the surface of the object must curve outward toward the observer.

Based on this definition, we propose dividing the femur into three convex substructures, each associated with a contour category, that are convex with respect to an observer (the C-arm) rotating around the diaphysis’s main axis: \textit{diaphysis}, the \textit{medial condyle} and the \textit{lateral condyle} (shown Figure \ref{fig:pipeline}) of class indices $I=\{1,2,3\}$ in the following.

Segmenting the femur CT model into these three substructures can be performed preoperatively. However, segmenting contour pixels for each substructure intraoperatively in the X-ray images presents several challenges.
The first challenge is the need for speed, as this process must be completed quickly to integrate seamlessly into standard surgical workflows. 
Additional challenges arise from the nature of medical imaging and the characteristics of X-ray images, such as low contrast, noise, overlapping structures, unclear or ambiguous boundaries, and artifacts.
For reference, see Figure~\ref{fig:training_data_synth}.

In response to these challenges, we employ the nnU-Net framework \cite{nnunet}, enhanced with Skeleton Loss \cite{skeletonloss}, to automatically and accurately segment the occluding contours of the diaphysis and condyles in X-ray images. nnU-Net is a self-configuring deep learning framework that automatically adapts its architecture and training parameters to the dataset. nnU-Net introduces key improvements, including dynamic adaptation of layers, feature maps, and hyperparameters to fit the input data. 
Skeleton Loss encourages the network to generate segmentation outputs with thin, connected structures corresponding to the object skeletons, in our case the contours, marking state-of-the-art performance for thin structure segmentation.
The total loss used for training denoted $\mathcal{L}_{\text{total}}$, is defined as a weighted sum of three components: the Dice loss $\mathcal{L}_{\text{dice}}$, 
 the Cross-Entropy loss $\mathcal{L}_{\text{CE}}$, and the Skeleton loss $\mathcal{L}_{\text{skeleton}}$, expressed as:

\begin{equation}
\mathcal{L}_{\text{total}} = \mathcal{L}_{\text{dice}} + \mathcal{L}_{\text{CE}} + \lambda \mathcal{L}_{\text{skeleton}}
\end{equation}
where $\lambda$ is a hyperparameter that adjusts the contribution of the Skeleton loss relative to the Dice and Cross-Entropy losses.
 The Skeleton Loss $\mathcal{L}_{\text{skel}}$, is defined as per the method in \cite{skeletonloss}. We employ Stochastic Gradient Descent (SGD) with Nesterov momentum as the optimizer, starting with an initial learning rate of 0.005 and applying weight decay and keeping the default of the nnU-Net implementation otherwise. The batch size is automatically adjusted by nnU-Net based on available GPU memory. The model is trained for 300 to 500 epochs for pretraining and 300 epochs for fine-tuning, with early stopping based on the validation loss values.

To train our model, we utilize a combination of pretraining and fine-tuning datasets. 
The pretraining dataset consists of DRRs \cite{Sherouse1990DRR} generated from 15 real CT scans of knee specimens. The femur is first segmented with submillimeter precision (see Online Resource 1, Section 4.2) from each CT scan by a medical professional using the Materialise Mimics software \cite{Mimics}.
The obtained 3D model is then divided into the medial condyle, lateral condyle, and diaphysis using a cut orthogonal to the first principal component at the center of mass, with the femoral head divided manually. Ground truth masks are created by projecting the contours of the 3D bone model onto the image plane, resulting in a mask-DRR training pair. In the fine-tuning step, we use specimen-specific CT data, which is typically acquired preoperatively in clinical practice. First, approximately 800 DRRs and corresponding masks are generated per specimen, covering angles from -90° to +90° relative to the ventral position, ensuring broad coverage.
Data augmentation is used to improve the model’s performance and reduce the domain gap between DRRs and real X-ray images. Augmentation techniques include contrast changes, Gaussian noise and image artifacts.
See Section 4 of Online Resource 1 for details. 

\subsection{Registration}
\label{sec:ICP}
Our multi-view X-ray/CT registration method consists of a contour-based ICP optimization. At each iteration of the ICP algorithm, the transformation minimizing the sum of the reprojection errors between the predicted contour points and the closest imaged CT point on the CT model silhouette is computed. The problem is stated as a non-linear least squares optimization problem, whose solution is written as:

\begin{equation}
\label{eq:BA_free}
\arg\min_{\{\mathbf{P}_c\}} \sum_j \sum_i \left\| \pi \left( \mathbf{P}_c, \mathbf{X}_{i}, \mathbf{K}_c \right) - \mathbf{x}_{j,i}\right\|_2^2 
\end{equation}
where $\mathbf{P}_c$ are the X-ray image pose parameters (parametrizing the rotation and translation), $\mathbf{K}_c$ the intrinsics (known) of the C-arm, $\mathbf{X}_{i}$ the 3D points of the bone substructure $i \in I$, $\mathbf{x}_{j,i}$ the 2D contour points of bone substructure $i$, and $\pi$ the projection function of the pinhole camera model. The optimization is carried out using the Levenberg-Marquardt algorithm implemented via the Python \texttt{scipy.optimize.least\_squares} function. The Jacobian of the residuals is computed numerically using finite difference approximations.

Once the optimal transformation is found, the 2D/3D correspondences for the three substructures are updated, and the next iteration begins. A key advantage of this method is that, unlike intensity-based methods, the residual error is interpretable, providing a clear indication of the registration quality.

Once the ICP has converged, meaning the matches are no longer updated, we apply correspondence reweighting based on the associated residuals and run the optimization a second time. The proposed reweighting discards points associated with large reprojection errors (e.g., points with residuals beyond double the standard deviation). This step has shown significant gains in the final registration accuracy.

\section{Experiments}
\label{sec:xp}
We describe the datasets used for evaluation in Section \ref{sec:dataset}, followed by the evaluation measures in Section \ref{sec:evalmet}, and the results in Section \ref{sec:results}.

\subsection{Datasets}
\label{sec:dataset}
Five cadaveric distal femur specimens (three female, two male) were obtained from a certified anatomical donation program with ethical approval. Four specimens were right-sided, and one was left-sided, with a BMI range of 19–28. Additional details are provided in Online Resource 1. The CT data were acquired using the Siemens CT NAEOTOM Alpha with a resolution of 512$\times$512px and a slice thickness of 0.2mm. The X-ray images were acquired using a Siemens XA - Fluorospot Compact S1 with a resolution of 976$\times$976px and a sensor size of 296.7$\times$296.7mm. 
During image acquisition, the specimens were placed on a radiolucent table, while the fiducial remained fixed relative to the bone throughout X-ray imaging.
The images were taken from 10 perspectives per specimen, with the C-arm rotated around the specimen at approximately 9-degree intervals. X-ray image poses were estimated using our fiducial marker.

\subsection{Evaluation Metrics}
\label{sec:evalmet}

We use the mean Reprojection Distance (mRPD) \cite{eval_1} as our evaluation metric, following the approach in \cite{jaganathan2022selfsupervised2d3dregistrationxray, Jaganathan_2021}
. 
MRPD represents the mean Euclidean distance between imaged 3D points and their corresponding observed 2D points in a control view, typically measured in millimeters.
For a set of 3D points $\{ \mathbf{X}_i \}_{i=1}^N$ and their images $\{ \mathbf{x}_i \}_{i=1}^N$, and the control view pose $\mathbf{P}_c$, the mRPD is defined as:

\[
\text{mRPD} = \frac{1}{N} \sum_{i=1}^{N} \| \mathbf{x}_i - \pi(\mathbf{X}_i; \mathbf{P}_c) \|_2
\]
where $\pi(\cdot; \mathbf{P}_c)$ 
is the projection function, 
mapping the 3D point $\mathbf{X}_i$ into the control view space, and $\| \cdot \|_2$ denotes the Euclidean norm. This metric is widely used in X-ray/CT registration \cite{eval_1, jaganathan2022selfsupervised2d3dregistrationxray}. In our experiments, the mRPD was calculated by comparing the ground truth control view bone contour, which was segmented by medical imaging specialists, with the reprojected bone contour after registration. 
Note that contours of bone substructures were not used in the calculation of the mRPD. An uncertainty analysis of ground truth segmentations is provided in Online Resource 1.

\subsection{Results}
\label{sec:results}

\noindent \textbf{Contour Extraction} Because precise location of the predicted contours is preferred over contour completeness in our registration method, we resort to the one-sided Chamfer distance from the predicted contours to the true contours to evaluate our contour segmentation model. Additionally, we report precision and recall. A predicted point is considered a true positive if it lies within a specified distance threshold of a true contour. Precision is defined as the proportion of these true positives relative to all predicted points, while recall is the proportion of true positives relative to all points on the true contours.
We achieved a one-sided Chamfer distance of \textbf{0.59}mm, a mean recall of \textbf{0.80} and a mean precision of \textbf{0.74} at a distance threshold of 1mm.

\begin{figure}[h]
    \centering
    \includegraphics[width=0.99\linewidth]{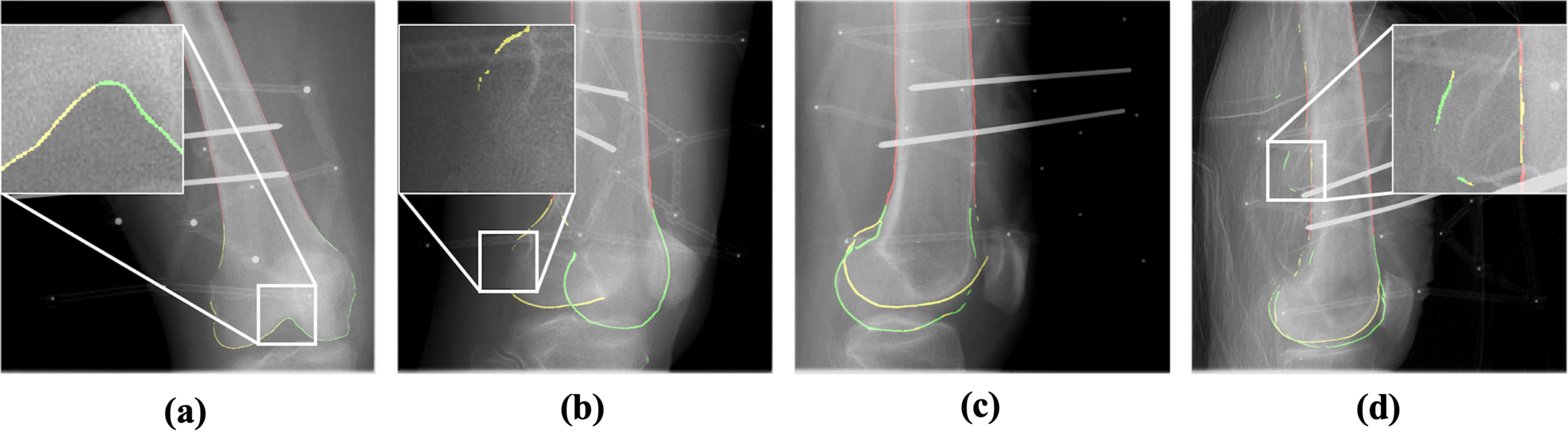}
    \caption{Example of bone contour predictions. Our model accurately segments the occluding contours of the bone substructures while distinguishing the medial condyle from the lateral condyle, even under challenging viewpoints, as shown in image \textbf{(c)}. Some images with strong noise can lead to false predictions as seen in image \textbf{(d)}}
    \label{fig:Example}
\end{figure}

\begin{figure}[h]
    \centering
    \includegraphics[width=0.99\linewidth]{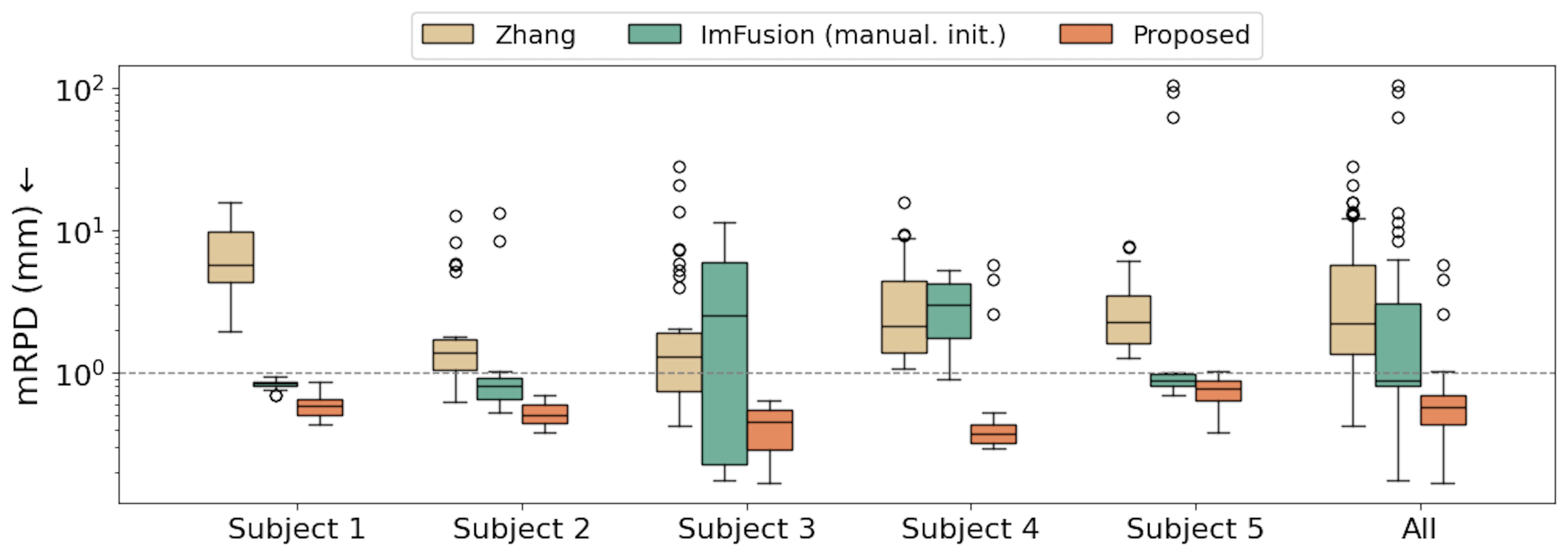}
    \caption{ Evaluation of different registration methods on our femur dataset via mRPD. A method successfully registered an X-ray image if the final mRPD was less than one millimeter (red line)}
    \label{fig:results_combined}
\end{figure}

\noindent \textbf{Registration} An evaluation of multiple registration strategies was performed. The commercial intensity-based method ImFusion \cite{ImFusion}, which is considered a leading approach in multi-view intensity-based registration. This method requires manual initialization to function effectively. Corresponding landmarks are manually placed by clicking correspondences in the images and CT data to achieve a rough alignment of the bone. Specifically, four corresponding points are manually marked in the images and CT using the provided GUI. These correspondences are then used to initialize the registration, bringing it close to the final solution. All registration parameters were kept at their default settings during this process. Additionally, we included a related ICP and contour-based approach from Zhang et al. \cite{zhang_contour}, with improvements by replacing Canny edge detection \cite{canny} with a deep-learning-based edge detection method \cite{bertasius2015deepedgemultiscalebifurcateddeep} in our implementation of the method. This method utilizes the bone silhouette, without any semantic substructures, during optimization. We did not find more open-source implementations of multi-view registration methods. Table~\ref{fig:results_combined} summarizes the registration errors on our real-world dataset. For each registration, we used two views with a perspective change of at least 45 degrees. Since our method does not require initialization, we set fixed values for all experiments, i.e., translations of +30mm, -40mm, +5mm and Euler angles of -17.18°, 0°, 17.18°. There are a total of 30 runs (6 runs per subject $\times$ 5 subjects) presented for each method. As described in Section \ref{sec:evalmet}, we use the mRPD as evaluation metric. The evaluation setup consisted of 3 control views, seperated by 9° each situated in between the two X-ray images used for registration. Our method achieved an mRPD of \textbf{0.67}mm, compared to \textbf{5.35}mm for the commercial method and \textbf{4.03}mm for Zhang's method. In some cases, our method produced incorrect registrations.
We demonstrate empirically in Online Resource 1 that these failing registrations can be automatically detected using the reprojection errors during the ICP optimization, allowing the registration to be restarted.
Note that this property holds under certain conditions regarding the quality of the contour segmentation results.

\noindent \textbf{Success Rate} Following the American Association of Physicists in Medicine (AAPM) guidelines \cite{eval_2}, a registration is defined as successful if the mRPD $\leq$ 1 mm. Our method achieved a success rate of 96.1\%, significantly higher than the 87.1\% success rate of the commercial solution. Additionally, Zhang's contour-based method showed much lower performance, with a success rate of 13.37\%.

\begin{figure}[h]
    \centering
    \includegraphics[width=0.99\linewidth]{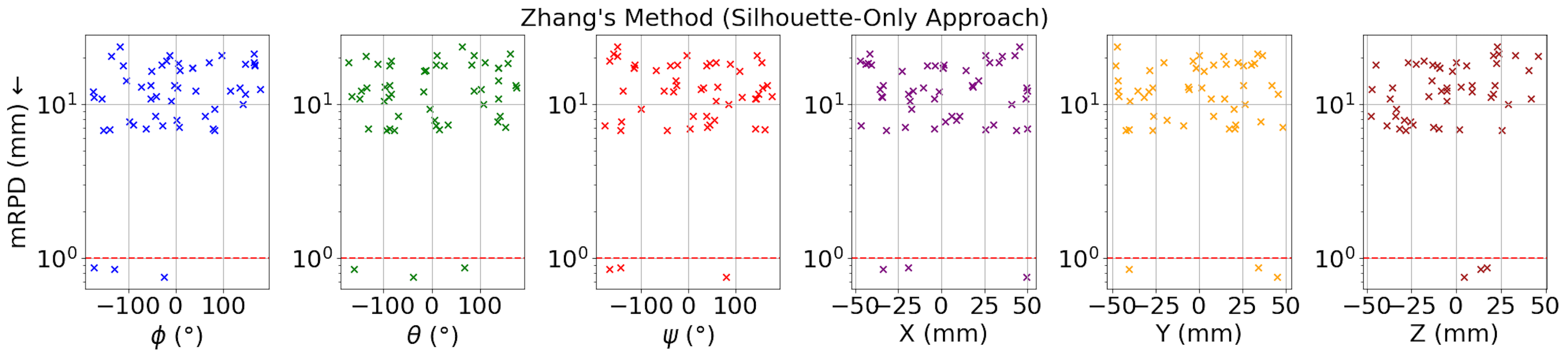}
    \includegraphics[width=0.99\linewidth]{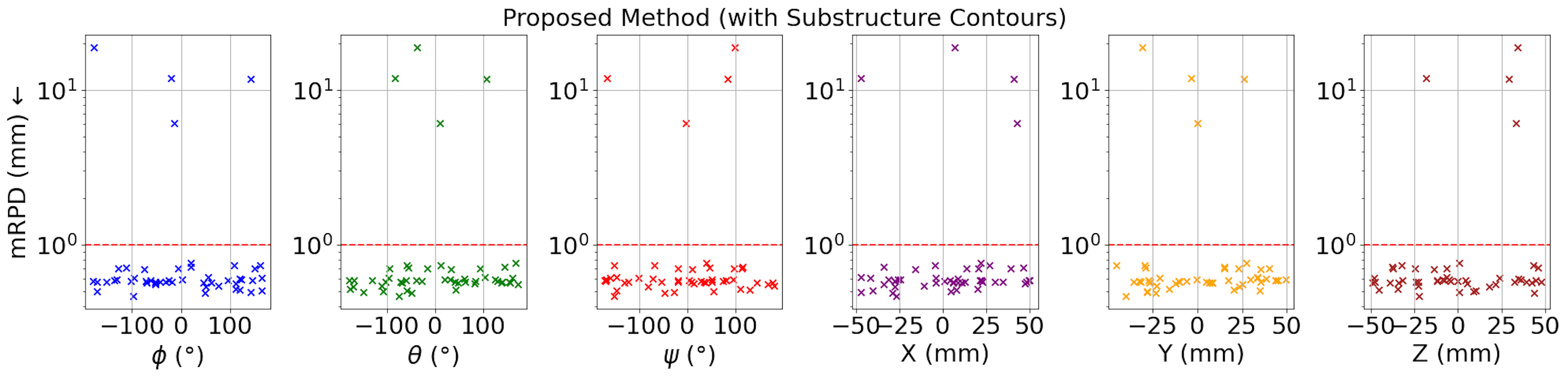}
    \caption{
    Robustness to random initialization of Zhang's method without substructures (top row) and the proposed method with substructures (bottom row).
  The abscissae represent initial parameter values, with each graph showing one sampled pose parameter. Each registration corresponds to six crosses with the same ordinate (mRPD) across graphs, representing the six initial pose dimensions}
    \label{fig:robust}
\end{figure}

\noindent \textbf{Robustness Analysis} To assess robustness of our method, we registered 50 scenes present in our dataset as seen in Figure \ref{fig:robust}, initializing Zhang's method and our method randomly within a range of -50mm to +50mm in each of the $(x, y, z)$ directions, along with a random initial rotation of -180° to +180° for each of the Euler angles $(\phi, \theta, \psi)$ relative to the registered position, covering the entire rotational space. 
Our method achieved 92\% submilimeter-accurate registrations across all specimens and initializations whereas Zhang's method, which does not make use of substructures, achieved only 8\%. When automatic detection of registration failures is incorporated in our method - triggering a perturbation when the final median ICP reprojection error exceeds 3mm - the method achieves a 100\% success rate across all experiments, requiring at most two runs per experiment and maintaining a final reprojection error of no more than 0.75mm across all experiments. Results with automatic restart are provided in Online Resource 1.

\noindent \textbf{Impact of Occluding Contours} A key innovation of our method is the use of component-level contours, as described in Section \ref{sec:method}. Their benefit in the ICP optimization is significant when compared to the object-level contours in Zhang's method, as shown in Figure \ref{fig:results_combined} and Figure \ref{fig:robust}. Zhang's method, which does not incorporate substructures, is more prone to getting trapped in local minima, as shown by the larger errors compared to our method.

\begin{figure}[h]
\centering
\includegraphics[width=0.7\linewidth]{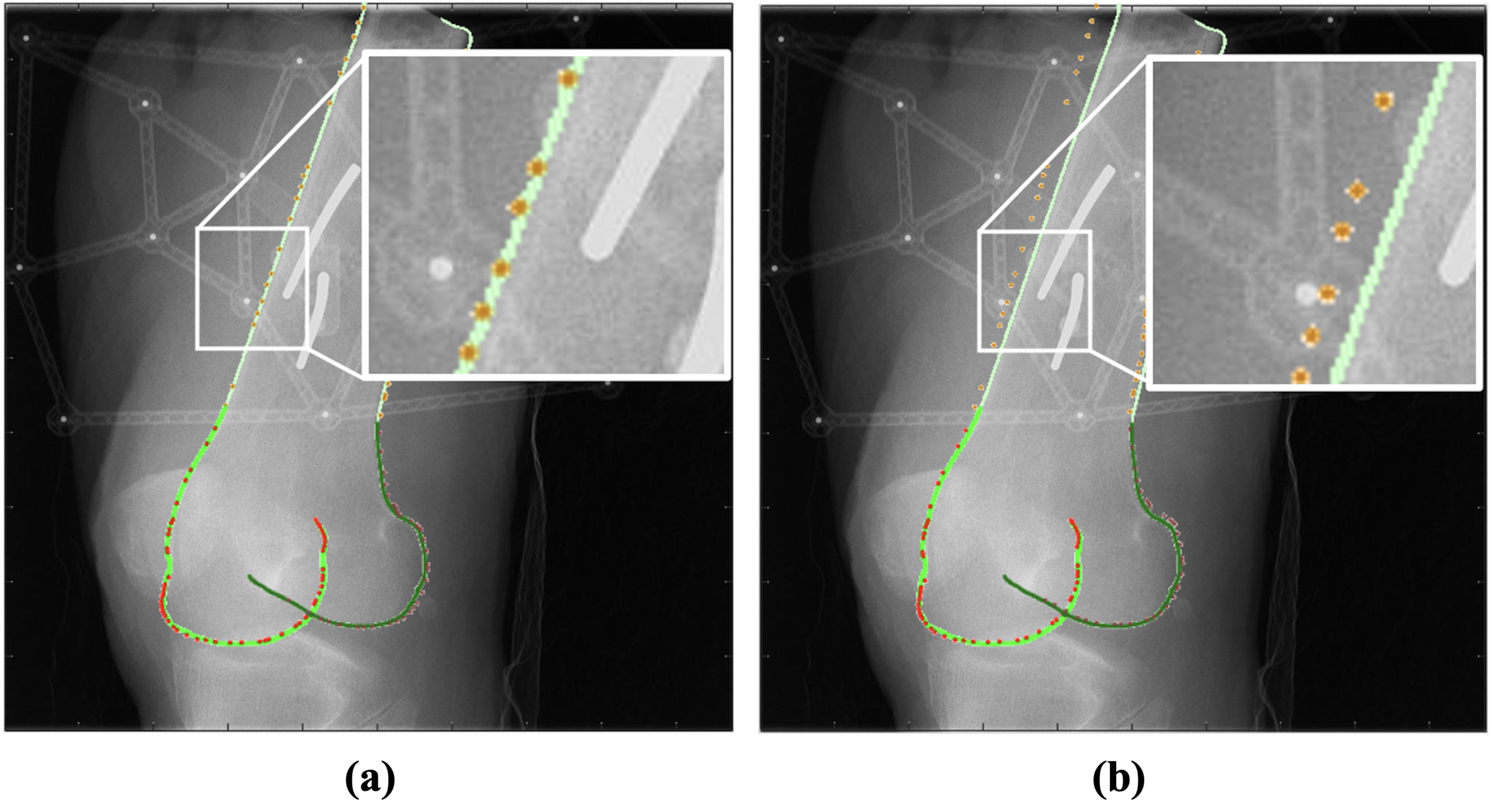}
\caption{This image shows a control view used for evaluation, with the ground truth contour in green and reprojected points from the decimated, registered bone mesh as colored dots. \textbf{(a)}: our method. \textbf{(b)}: ImFusion}
\label{fig6}
\end{figure}

\noindent \textbf{Visual Results} Figure~\ref{fig6} shows a qualitative comparison of our method with ImFusion. Green contours represent the ground truth, and the colored dots show reprojected points from the registered decimated bone mesh.

\section{Discussion}
\label{sec:conclusion}
\noindent \textbf{Limitations and Future Work}
A key limitation of our method, which is not present in intensity-based methods, is its reliance on the accuracy of the CT segmentation.
Our method has shown strong performance on the femur bone. A preliminary experiment on a humerus has been performed (see Online Resource 1, Section 3.6) and suggests that our method generalizes well to other anatomical structures. Future work will focus on generalizing the approach to other anatomies, including those with more complex shapes. 
While CT segmentation into the three proposed substructures is relatively straightforward in the case of the femur, future work will focus on transitioning from an ad hoc solution tailored to the femur to a more generic approach that can be applied across various anatomies.
Integrating the registration method into surgical navigation systems is also a key focus.
Additionally, the testing of our method was conducted using data from a single C-arm device. A more extensive study is needed to evaluate the generalizability of the method to other imaging devices and a wider range of real-world imaging conditions.

\noindent \textbf{Conclusion}
In this work, we introduced a novel, fully automated multi-view X-ray/CT registration method that leverages bone substructure contours to achieve accurate and robust intraoperative registration for orthopedic surgeries. By focusing on matching specific contours of bone substructures, the diaphysis and condyles, we reduced the ambiguity inherent in traditional silhouette-based methods. This approach not only enhances the robustness of the Iterative Closest Point (ICP) optimization but also significantly improves registration accuracy compared to a commercial solution requiring manual key-point annotation.

\section*{Acknowledgments}
This work has been funded by the Wonderland Foundation and supported by the Swiss Center for Musculoskeletal Imaging (SCMI), OR-X - a Swiss national research infrastructure for translational surgery - and associated funding by the University of Zurich and University Hospital Balgrist.

\section*{Declarations}

\noindent \textbf{Competing Interests} The authors have no relevant financial or non-financial interests to disclose.

\noindent \textbf{Compliance with Ethical Standards}
All procedures performed in studies involving human specimens were in accordance with the Swiss national ethical standards. The Kantonale Ethikkommission Zurich approved the study.

\noindent \textbf{Informed Consent}
This study did not involve live human participants, and therefore, informed consent was not required.

\begin{appendices}
{This supplementary material provides technicalities, additional experiments, detailed analysis of the experiments and further information regarding dataset and model training. We provide the reader with an explanation of the C-Arm calibration in Section \ref{sec:calib}. We then provide additional results, including ablation studies in Section \ref{sec:add_results}. In Section \ref{sec:dataset}, we present further details about our dataset and qualitative analysis of the ground truth segmentations. Finally, we give details about our model training in Section \ref{sec:impl_details} and more information about the memory usage of our method in Section \ref{sec:memory}.}

\section{C-Arm Calibration}
\label{sec:calib}

A prerequisite to multi-view X-ray/CT registration is C-arm calibration, which refers to the computation of the relative poses of the X-ray images. Our C-arm calibration is achieved using a fiducial marker, similar to prior works \cite{goldstandart, multi_view_registration_2011}. {This approach offers a simple, effective, and accurate solution to this problem.} We utilize a custom-designed fiducial marker, which is mounted on a K-wire attached to the femur at the start of the surgery. After the X-ray images are acquired, the marker can be removed. This procedure also offers the possibility of extending the registration system to optical or EMT-based tracking systems \cite{Franz_EMT}, which can be calibrated with the phantom prior to the surgery.
For procedures like osteotomies, where a K-wire is already placed on the bone, our method only minimally deviates from the standard clinical workflow. 

\begin{figure}[h]
    \centering
    \centering
    \includegraphics[width=0.95\linewidth]{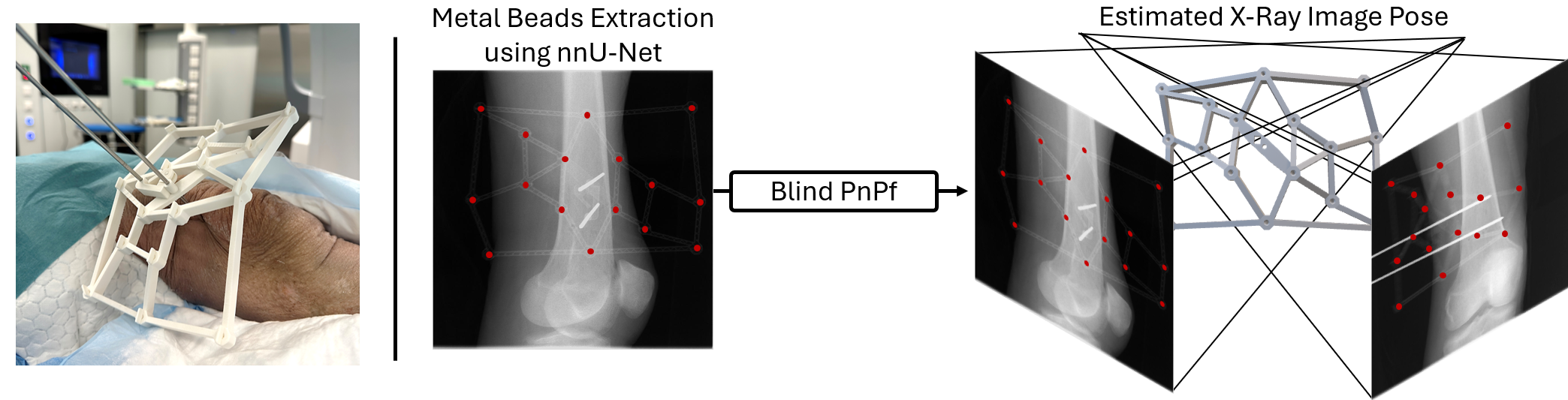}
    \caption{Left: 3D-printed fiducial attached to the distal femur of a specimen. Right: C-arm calibration pipeline. Beads are automatically segmented, and their centers are used to solve a blind PnP, generating 2D-3D correspondences. These correspondences are then used to compute the final X-ray images poses}
    \label{fig:bead_structure}
\end{figure}

Our custom-designed fiducial marker is 3D-printed using radiolucent material, with 16 embedded metal spheres of 2 mm diameter. The design was inspired by previous works \cite{multi_view_registration_2011}. The choice of the number of beads and their spatial configuration was based on preliminary synthetic simulations. A spatial configuration that offers a good trade-off between bead self-occlusion, field of view-induced occlusion, and calibration accuracy was selected. This results in a non-coplanar configuration of beads distributed across a large portion of the C-arm’s field of view (FOV).
The clamping system for fixation on the K-wire uses no metal parts, reducing artifacts in the X-ray images. The design is illustrated in Figure \ref{fig:bead_structure}.

The calibration procedure is performed as follows. The 2D positions of the metal spheres are segmented in the X-ray images using a trained nnU-Net \cite{nnunet}.
They are then matched with their known 3D positions. 
The matching is performed through an exhaustive search over all possible subsets of 2D-3D correspondences associated to minimal solutions to the Perspective-n-Point (PnP) problem under unknown focal length while assuming a pinhole camera model \cite{pinhole_2}.
The matching solution maximizing the number of inliers, namely 2D-3D matches with low reprojection errors (lower than 0.8 pixels or 0.26mm in our experiments), is retained.
The final X-ray image poses, denoted as ${\mathbf{P}_c}$ {in the main paper}, are computed using these inliers.

\begin{figure}[h!]
    \centering\includegraphics[width=0.78\linewidth]{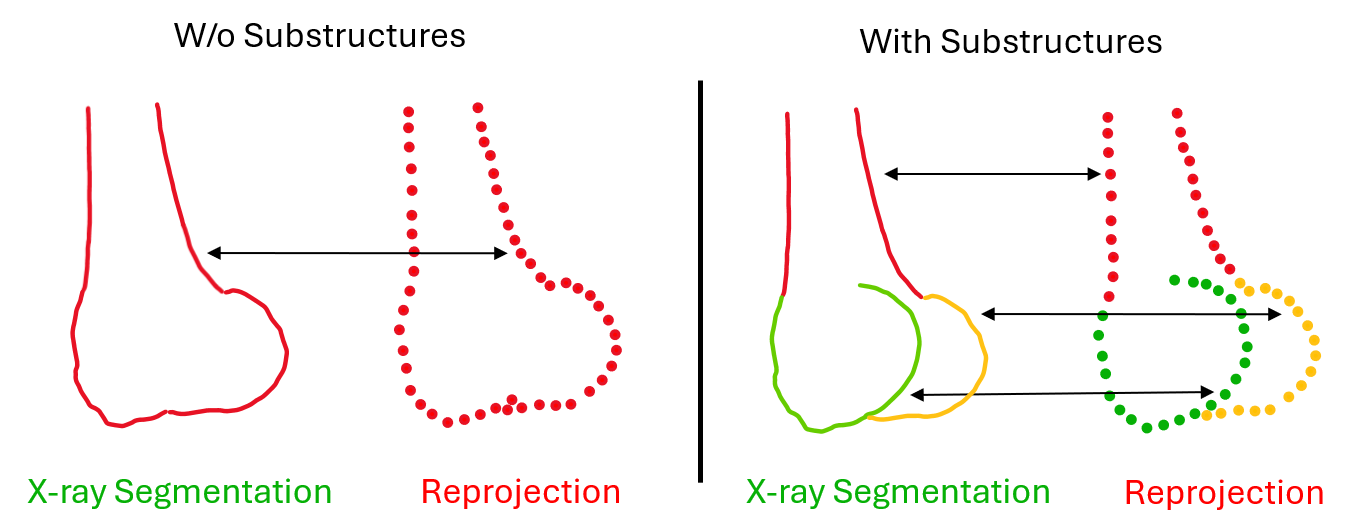}
    \caption{{Overview of point matching in ICP.
Left: Points are matched along the occluding contour of the entire femur, specifically over its silhouette (Zhang's method \cite{zhang_contour}). The associated registration results are referred to as 'Zhang's Method (Silhouette-Only Approach)' in Figure 5 of the main paper.
Right: Proposed method in which points are matched within the same class of substructures}}
    \label{fig:substructures}
\end{figure}

{\section{Matching Principle}}
{In this section we want to give a visual explanation of our contour matching. Unlike registration methods relying on object-level contours (see Figure \ref{fig:substructures}, left), we propose extracting component-level contours, specifically those of anatomical substructures (see Figure \ref{fig:substructures}, right).}

{\section{Additional Results}
\label{sec:add_results}}
{\subsection{Ablations}}

\begin{figure}[h]
    \centering
    \includegraphics[width=0.99\linewidth]{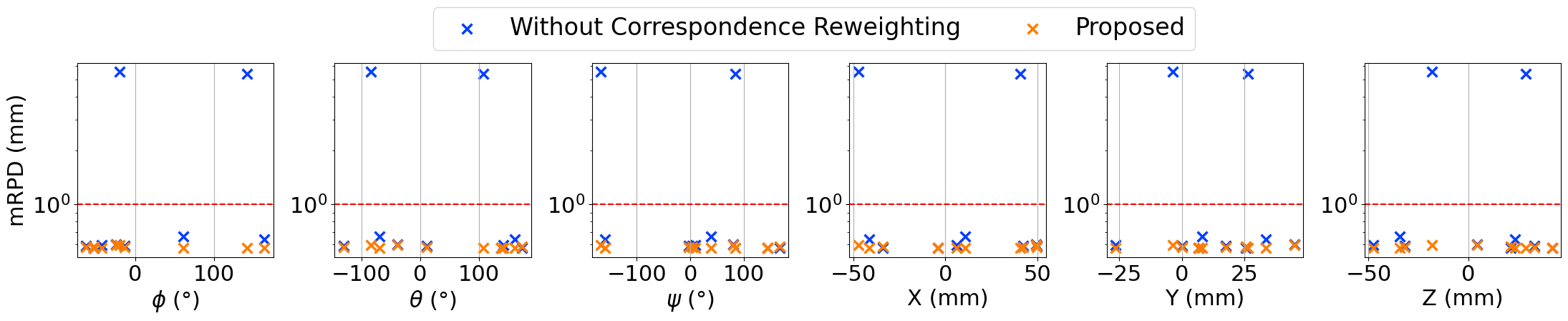}
    \caption{{mRPDs for 10 random initializations of our method with and without correspondence reweighting are shown. Note that some crosses may overlap. Without correspondence reweighting, the mean mRPD is 1.574mm, whereas our proposed method achieves a mean mRPD of 0.576mm}}
    \label{fig:ablations}
\end{figure}

{We propose an ablation study to evaluate the impact of correspondence reweighting on the robustness and accuracy of the registration. This reweighting process excludes points with large reprojection errors in the proposed ICP optimization.
The mRPDs for 10 randomly initialized registrations (with a fixed seed) on the data from one subject are shown in Figure \ref{fig:ablations}, and the corresponding mean mRPDs are 1.574mm and with our proposed method we achieve 0.576mm. Correspondence reweighting contributes to improved robustness and accuracy, resulting in approximately 1mm of improvement in mRPD.}

\begin{figure}[h]
    \centering
    \begin{subfigure}[b]{0.325\linewidth}
        \centering
        \includegraphics[width=\linewidth]{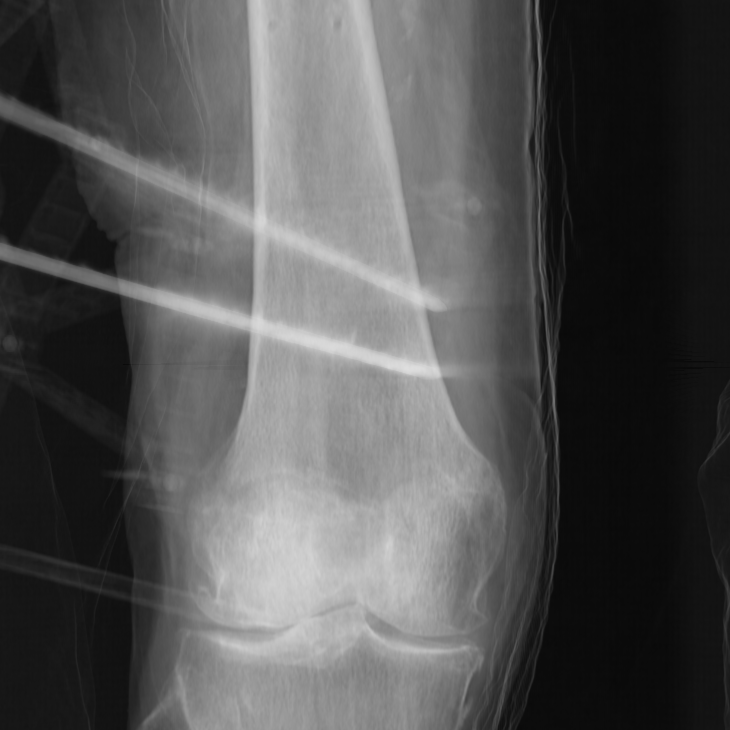}
    \end{subfigure}
    \begin{subfigure}[b]{0.325\linewidth}
        \centering
        \includegraphics[width=\linewidth]{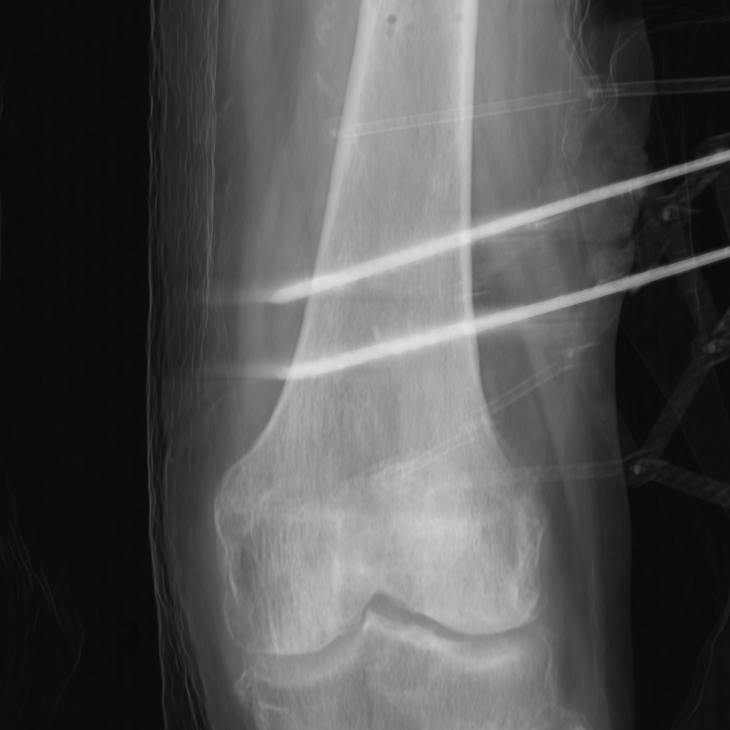}
    \end{subfigure}
    \begin{subfigure}[b]{0.325\linewidth}
    \includegraphics[width=\linewidth]{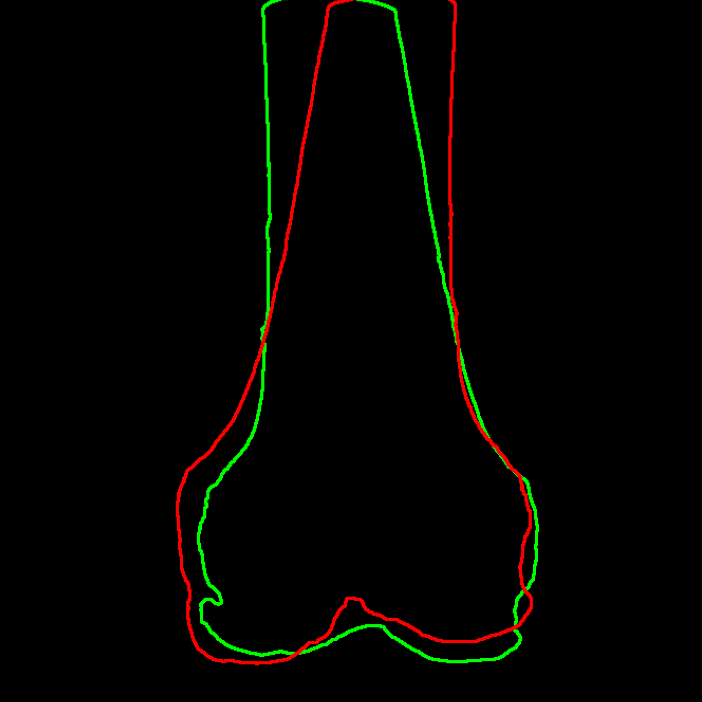}
    \end{subfigure}
    \centering
    \caption{Illustration of ambiguity in single-view registration of symmetrical bones. Left: X-ray acquired from ventral to dorsal. Middle: X-ray acquired from dorsal to ventral. Right: overlay of their contours, highlighting the high degree of similarity between these X-rays}
    \label{fig:ambiguity}
\end{figure}

\begin{figure}[h]
    \centering
    \includegraphics[width=1.0\linewidth]{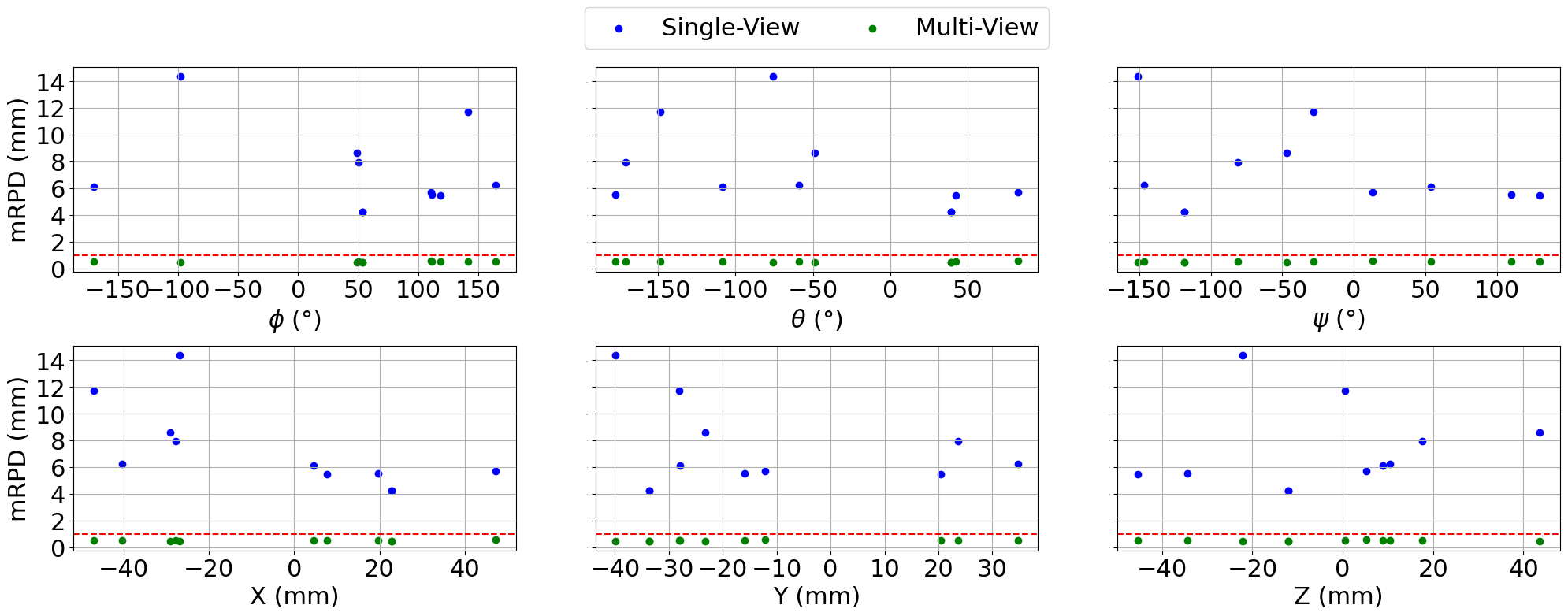}
    \caption{mRPDs resulting from single-view and multi-view registrations with our method over 10 runs (performed on a single subject). The multi-view approach consistently outperforms the single-view approach}
    \label{fig:single_vs_multi}
\end{figure}

{\subsection{Single- vs. Multi-View}}
\label{subsec:singlevsmulti}
{As highlighted in prior research \cite{grupp2020automatic, multi_view_registration_2011, point_net}, single-view registration is significantly less reliable than multi-view approaches, particularly for symmetric objects like the femur. In Figure \ref{fig:ambiguity}, we provided an example of different views with strong similarities due to the symmetries of the femur. We provide additional results to support these findings, including a quantitative comparison of single-view versus multi-view performance of our method in Figure \ref{fig:single_vs_multi}. We conducted 10 registrations in the experiment with random initializations (using a fixed seed) for the same subject, employing identical views. The multi-view approach consistently achieves submilimetric accuracy, while the single-view approach performs poorly with no registration showing a mRPD below 1mm.}

\newpage

{\subsection{Robustness}\label{subsec:robustness}}

{
In this section, we present additional results supporting the claim that the interpretability of the residual error in the proposed ICP optimization can be leveraged to detect registration failures, therefore allowing for the automatic restart of the optimization.}

\begin{figure}[h!]
    \centering
    \includegraphics[width=0.99\linewidth]{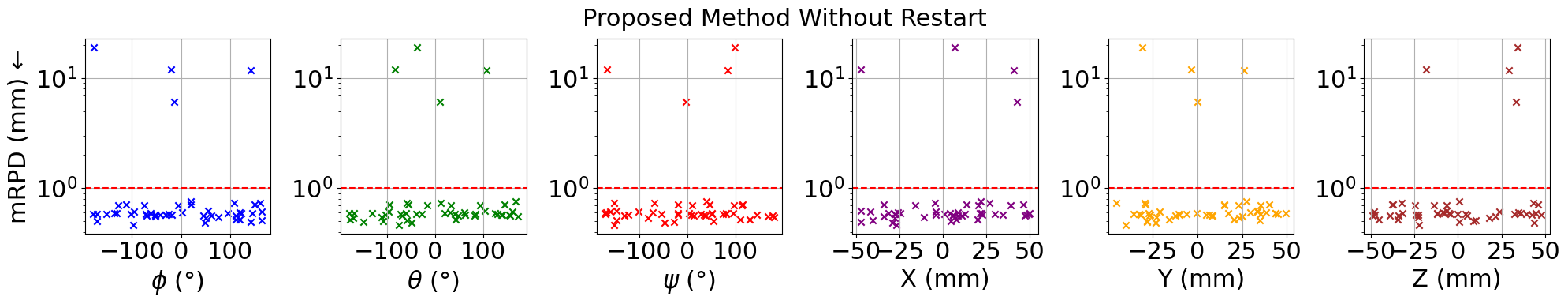}    \includegraphics[width=0.99\linewidth]{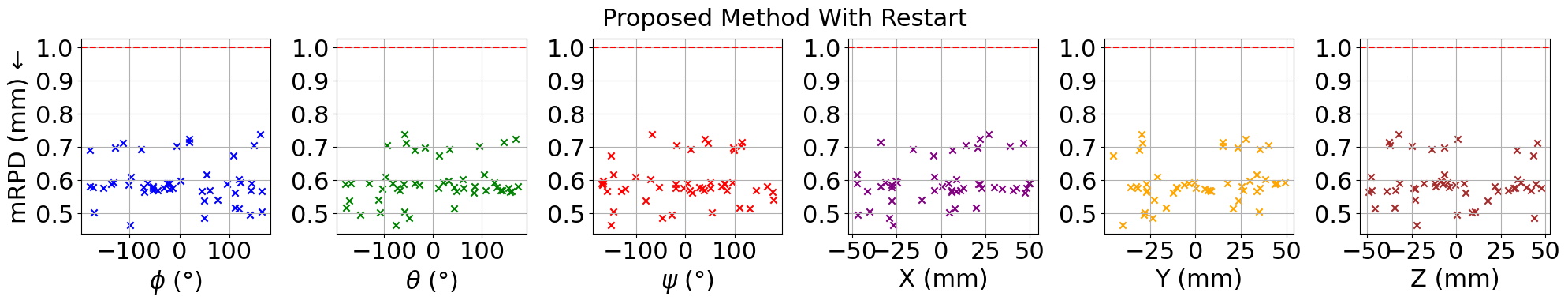}
    \caption{{Each plot shows 50 data points, each representing the mRPD of a randomly initialized run. Top: our method without restart (as detailed in the main article). Bottom: the enhanced method with restart. The x-axis indicates the initial perturbation of the corresponding parameter. Overall, we achieved a 100\% success rate with a maximum mRPD of 0.75 mm}}
    \label{fig:multi_registration}
\end{figure}

{The proposed algorithm is as follows.
If, after the first 4 correspondence updates in our ICP optimization, the median reprojection error remains higher than a predefined threshold (empirically set to 3mm), 
the ICP optimization is restarted. This restart involves applying a random perturbation, selected within the range $[90^\circ, 270^\circ]$, to the previous value of the $\psi$ parameter, which represents a rotation around the diaphysis's main axis. In our experiment, where we randomly initialized 50 registrations across all subjects, the few runs that previously ended in a local minimum now also converge. Results are shown in Figure \ref{fig:multi_registration}. This experiment achieved a 100\% success rate, with random initialization from -180° to +180° for each angle. The results show an average mRPD of 0.59 mm, with a maximum of 0.75 mm and a standard deviation of 0.07 mm. Only 4 out of 50 runs required a restart.}

{The range of precision and recall requirements that the proposed contour segmentation model must meet for the above algorithm to be applicable has not been studied and is left for future work.}

\begin{figure}[h]
    \centering
    \begin{subfigure}{0.32\linewidth}
        \centering
        \includegraphics[width=\linewidth]{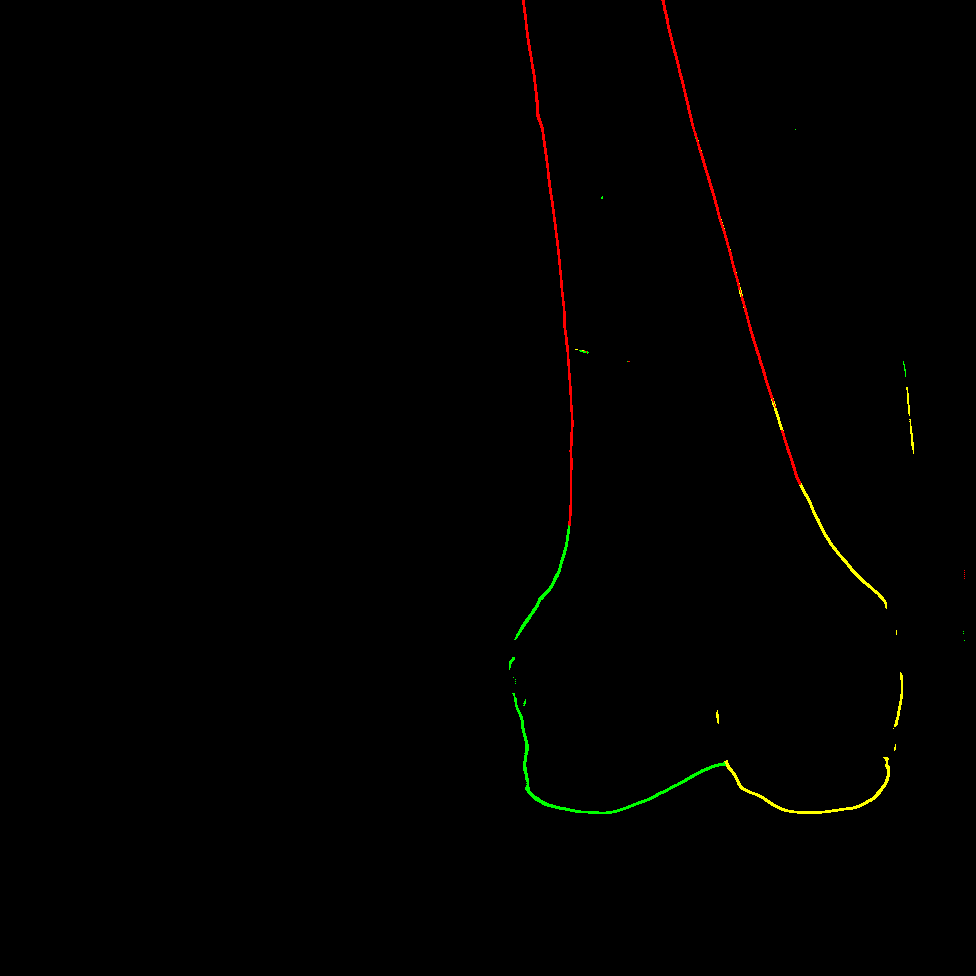}
    \end{subfigure}
    \begin{subfigure}{0.32\linewidth}
       \centering
        \includegraphics[width=\linewidth]{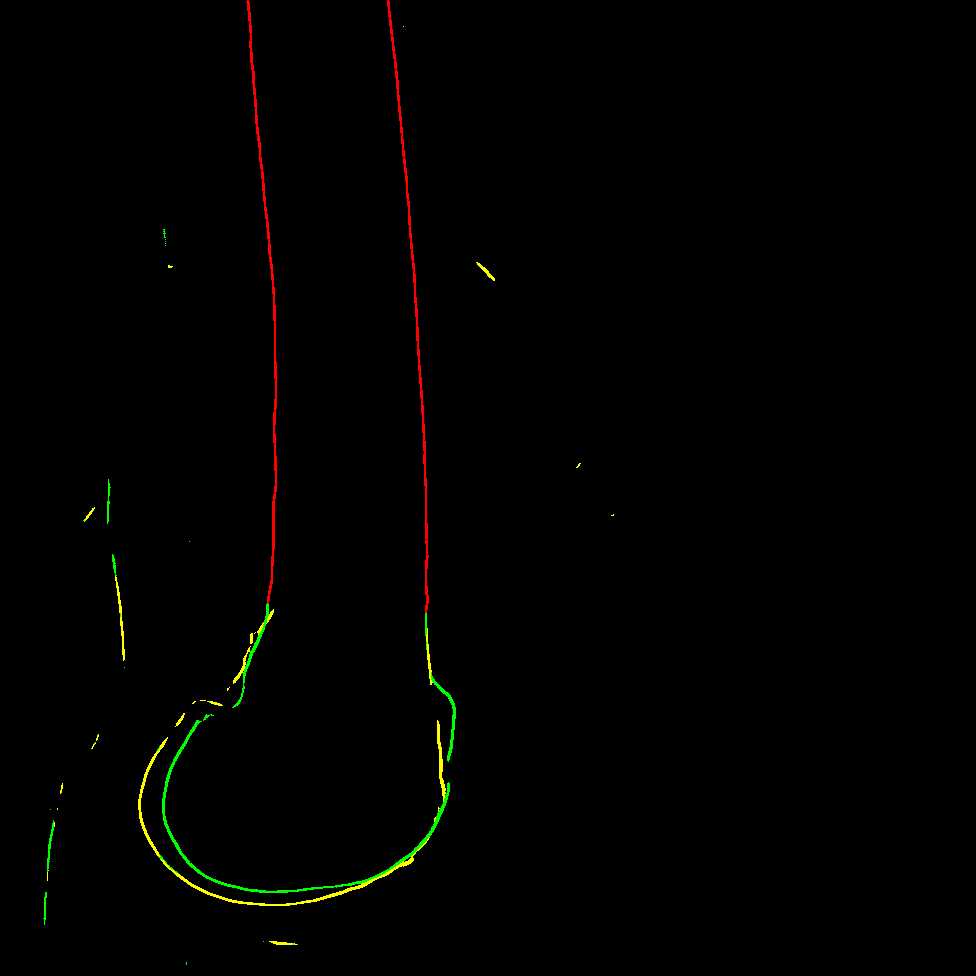}
    \end{subfigure}
        \begin{subfigure}{0.32\linewidth}
       \centering
        \includegraphics[width=\linewidth]{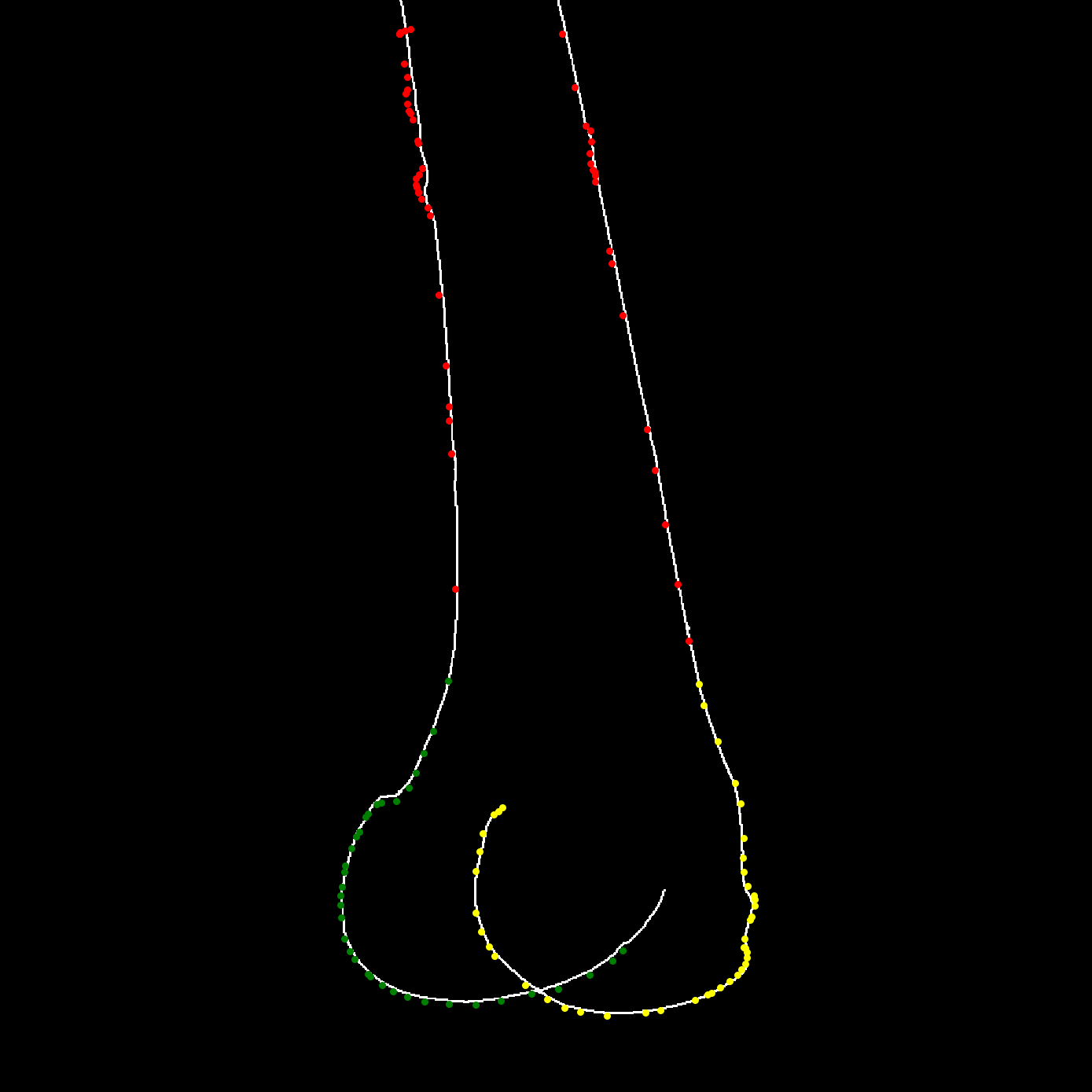}
    \end{subfigure}
    \caption{{Qualitative example showing the registration on a multi-view example with a substantial amount of false and misclassified predictions. Left and middle: the predicted X-ray image contours. Right: the reprojected registered bone contour overlaid on a control view. With a random initialization, this image pair leads to a successful registration of 0.71mm mRPD}}
    \label{fig:noisy_regist}
\end{figure}

{\subsection{Robustness to Noisy Contour Prediction}}
{In this section, we provide a qualitative evaluation of the performance of the registration in the presence of false and misclassified contour predictions. Figure \ref{fig:noisy_regist} shows contours extracted from X-ray images. These contours include significant false predictions, well visible on the image in the middle. After running our method, starting from a random initialization within -180° to +180° in each of the Euler angles and -50 mm and +50 mm in each of the Cartesian coordinates, we achieve an mRPD of 0.71 mm. This example illustrates the robustness of our method in successfully registering the bone model to contour predictions with a substantial amount of false and misclassified predictions.}

{\subsection{Statistical Significance}
We evaluate the statistical significance of our results reported in Figure 4 of the main paper using a non-parametric Friedman test across all specimens. Specifically, we analyze the significance of the registration accuracy across 5 specimens with the same deterministic initialization and view combinations. Our method demonstrated statistically significant improvements over Zhang's method across all subjects (p $<$ 0.05) and outperformed ImFusion in most cases (p $<$ 0.05 for subjects 1, 2, 4, and 5). The Friedman test confirmed overall significance between methods (p $<$ 0.001) for all specimens, except for subject 3, which showed high uncertainty. This uncertainty prevents us from making definitive claims about statistical significance for this subject, despite our method showing a clear improvement in the mean performance.}

\begin{figure}[h!]
\centering
    \begin{subfigure}{0.32\linewidth}
        \centering
        \includegraphics[width=\linewidth]{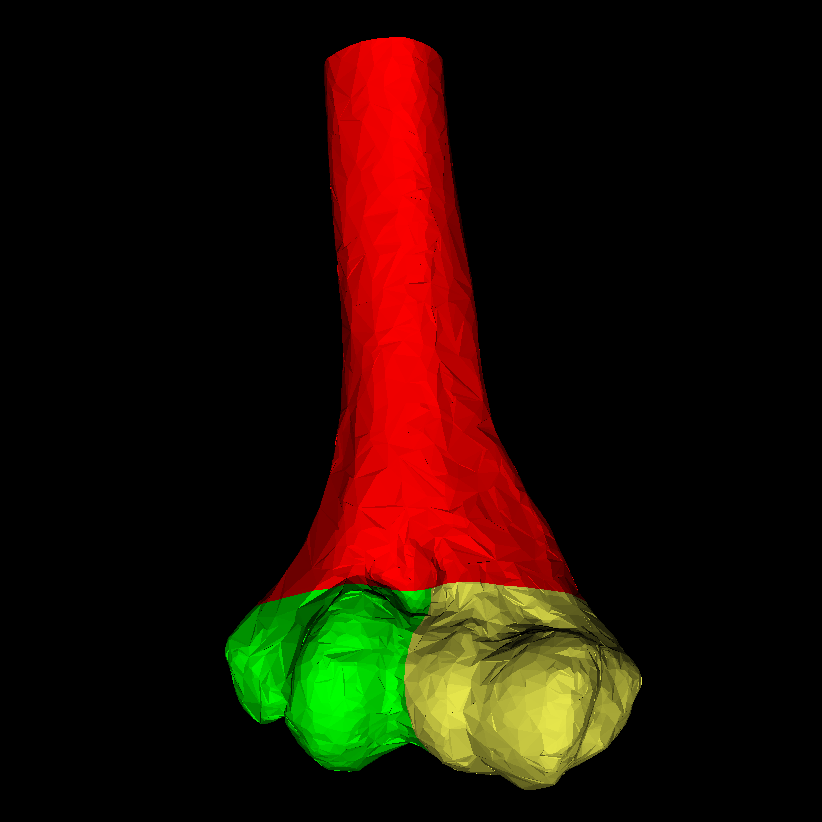}
    \end{subfigure}
    \hspace{-2mm} 
        \begin{subfigure}{0.32\linewidth}
        \centering
        \includegraphics[width=\linewidth]{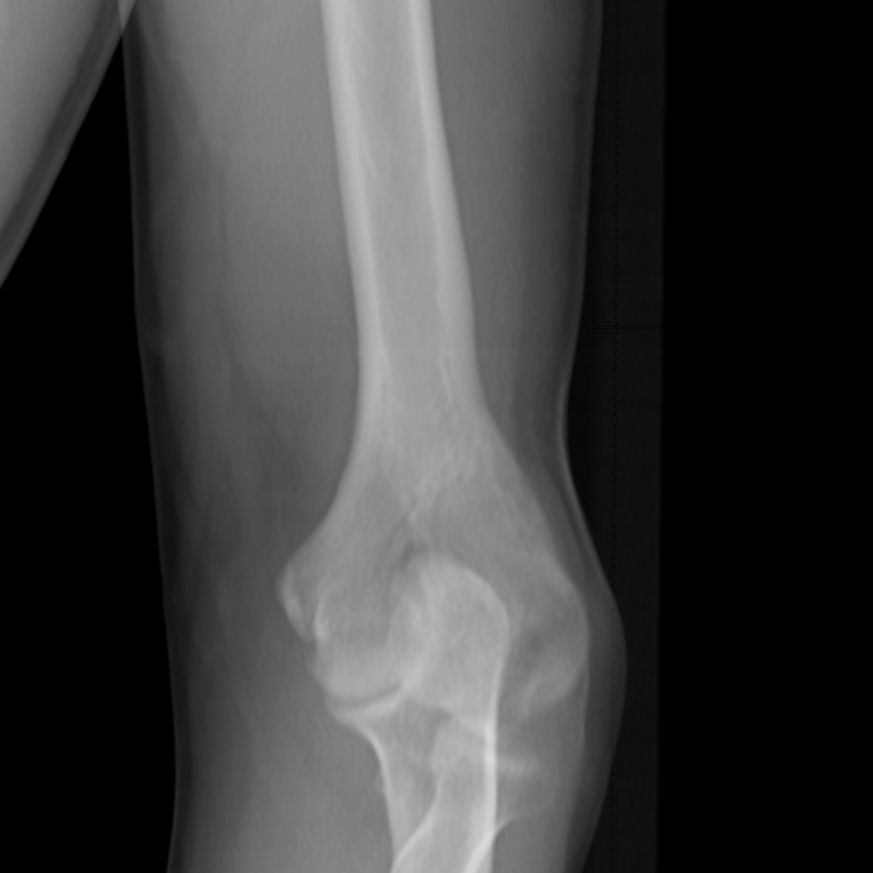}
    \end{subfigure}
    \hspace{-2mm} 
    \caption{{Left: Image showing the slicing of the distal humerus in 3 parts, analogous to the slicing performed with the femur. Right: DRR's from a random view}}
\label{fig:sliced}
\end{figure}

{\subsection{Generalization}
In the main paper, our experiments focused exclusively on the femur. To evaluate the applicability of our method to different anatomies, we conducted an additional experiment using the humerus. The data for this experiment were derived from a CT scan of a 41-year-old male patient's left arm, obtained with general consent. The experimental setup involved generating synthetic X-ray data from various perspectives and extracting contours following the same methodology as in the femur experiments.
The humerus was segmented into three pieces, similar to the femur, as illustrated in Figure \ref{fig:sliced}.}

\begin{figure}[h!]
\centering
    \includegraphics[width=0.95\linewidth]{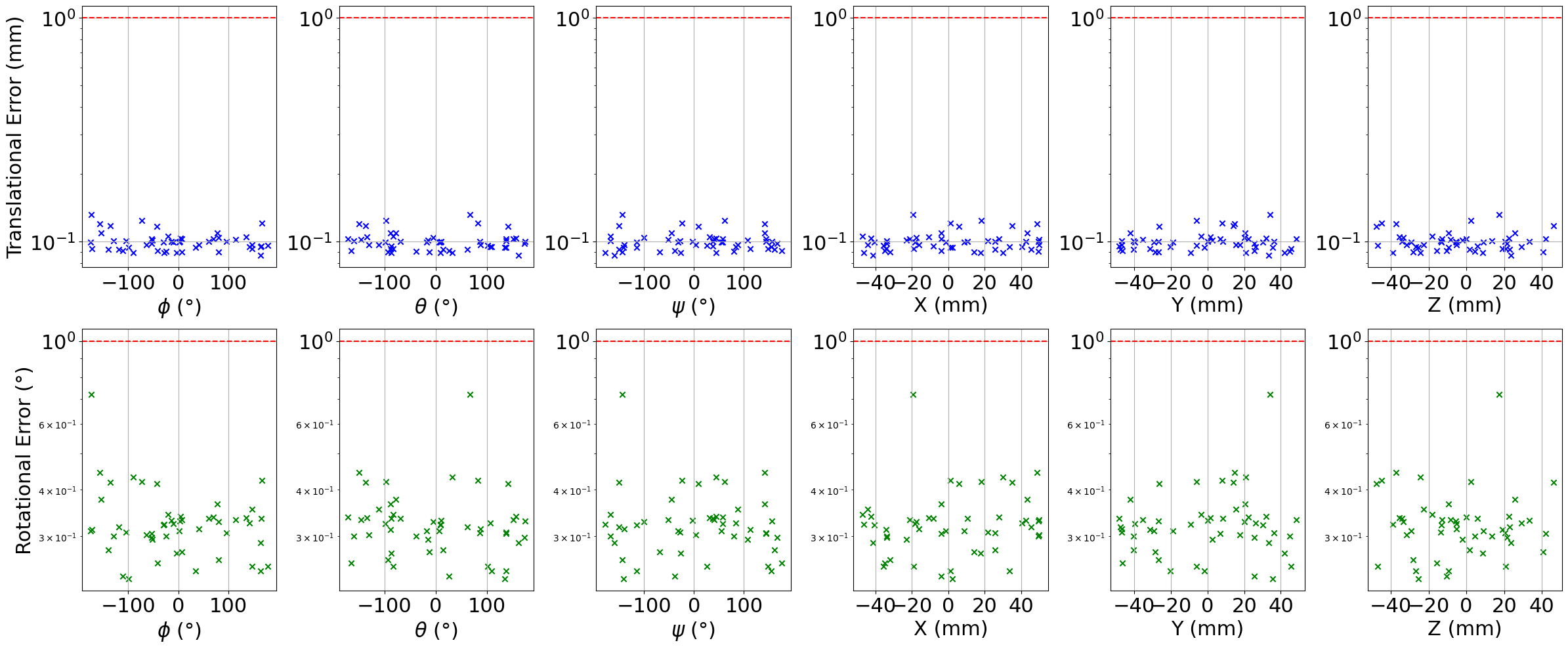}
    \caption{{Each plot shows 50 data points, each representing the error of a randomly initialized run. The x-axis indicates the initial perturbation of the corresponding parameter. This experiment achieved a 100\% success rate, with a maximum translational error of 0.13 mm and a maximum angular error of 0.72°}}
\label{fig:humerus_regist}
\end{figure}

{Similar to the robustness experiments detailed in the main article (see Figure 5), we conducted 50 registrations with random initialization using two fixed views. The results, shown in Figure \ref{fig:humerus_regist}, demonstrate a 100\% success rate with translational errors below 1 mm. As anticipated, the errors are smaller compared to our other experiments, owing to the use of synthetic data rather than real-world X-rays. These results suggest that our method has the potential to generalize effectively to other long bones and to more complex anatomical structures. However, further validation using real-world data is essential to substantiate these claims and will be addressed in future work.}

{\section{Dataset}
\label{sec:dataset}
\subsection{Details}
The compiled dataset includes 5 specimens. Table \ref{tab:dataset} provides the height (cm), weight (kg), BMI, age (years), sex and side of each specimen.}

\begin{table}[h!]
\centering
\caption{{Specimen Characteristics}}
\begin{tabular}{ccccccc}
\hline
{Height (cm)} & {Weight (kg)} & {BMI} & {Age (years)} & {Sex} & {Side} \\
\hline
{185}          & {79}           & {23}          & {75}              & {Male}   & {Right}\\
{183}          & {65}           & {19}          & {71}              & {Male}   & {Right}\\
{155}          & {63}           & {26}          & {76}              & {Female} & {Right}\\
{157}          & {61}           & {25}          & {65}              & {Female} & {Left}\\
{160}          & {71}           & {28}          & {66}              & {Female} & {Right}\\
\hline
\end{tabular}
\label{tab:dataset}
\end{table}

\begin{figure}[h]
    \centering
    \includegraphics[width=0.98\linewidth]{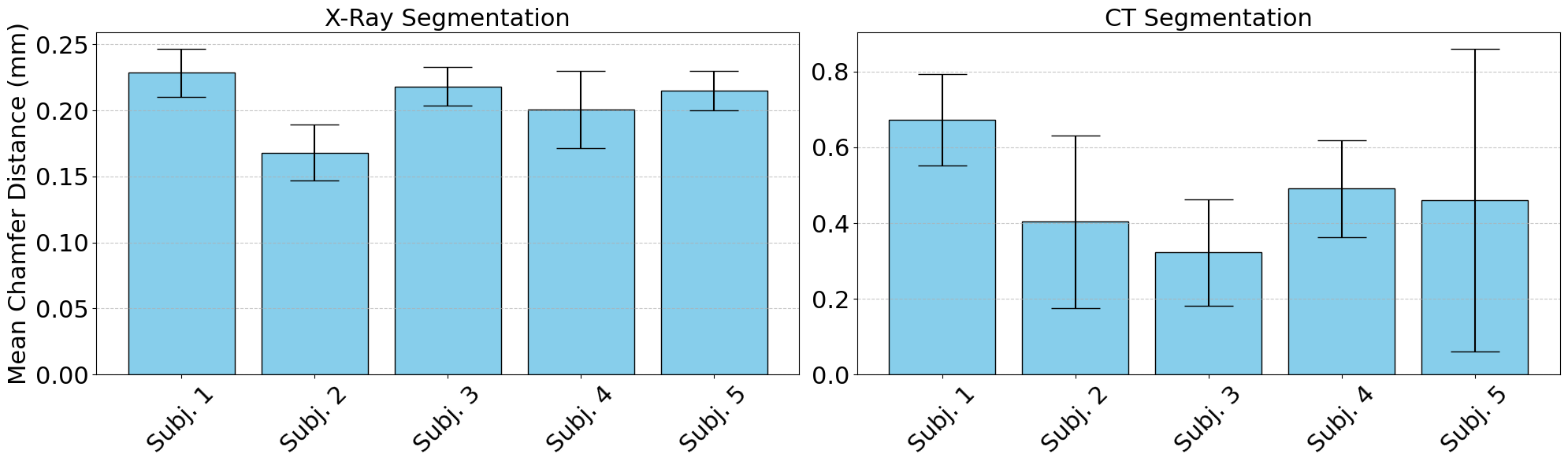}
    \caption{{Pairwise Chamfer Distance for CT and X-Ray Segmentations. For each specimen, we report the mean Chamfer distance of pairwise combinations of segmentations across all annotators for that subject, along with the standard deviation. Left: Results for X-ray contour segmentations. Right: Results for CT segmentations}}
    \label{fig:segmentation1}
\end{figure}

{
\subsection{Ground Truth Analysis}
Our dataset needs two types of manual or semi-automatic annotated ground truth data: the CT segmentation of the femur and the control-view X-ray segmentation. These tasks are performed by medical professionals who specialize in providing these services for surgical planning. To give a more detailed analysis of the segmentations, we have asked 4 of them to segment the same X-ray images and CT scans. In the main paper, the different segmentations were performed by different annotators. Figure \ref{fig:segmentation1} presents the uncertainty analysis for both types of segmentation. For each specimen, we report the mean Chamfer distance across all pairwise combinations of segmentations (e.g., annotator 1 vs. annotator 2, annotator 1 vs. annotator 3, etc.) along with the standard deviation. 
The mean Chamfer distances are 0.47 mm for CT segmentations and 0.21 mm for X-ray segmentations across all annotators and specimens. This numerical evaluation shows a submilimetric segmentation precision.}

\newpage

{\section{Implementation Details}
\label{sec:impl_details}}
{\subsection{Training Details}}
\begin{figure}[h]
    \centering
    \includegraphics[width=0.9\linewidth]{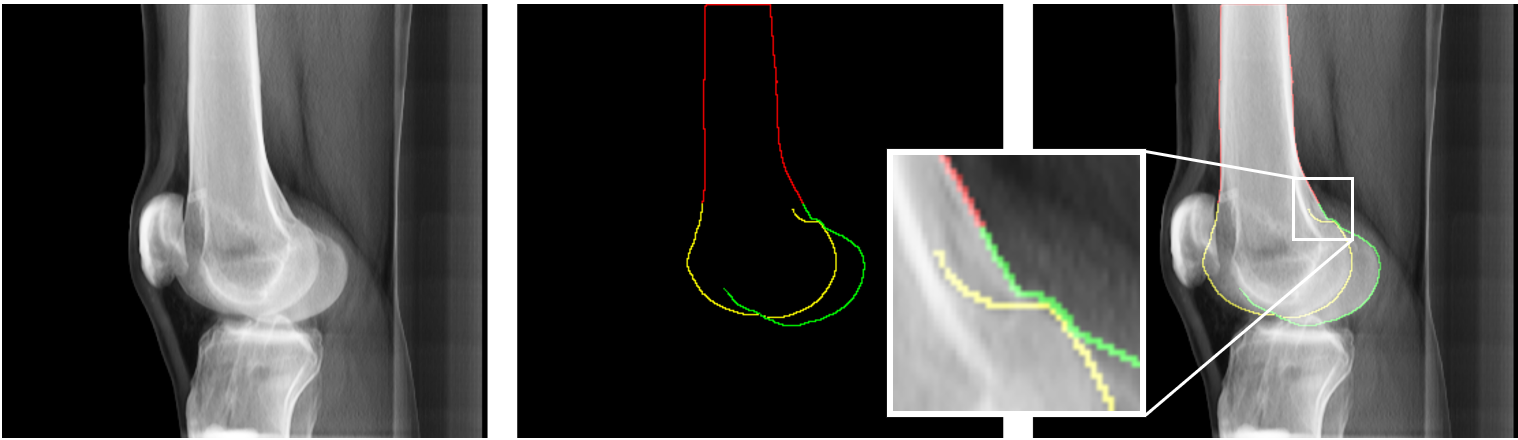}
    \caption{Example of a training sample. Left: DRR. Middle: Contour labels of the femur diaphysis (red), medial and lateral condyles (green and yellow resp.). Right: Overlay of both}
    \label{fig:training_data_synth}
\end{figure}

\begin{figure}[h]
    \centering
    \begin{subfigure}{0.225\linewidth}
        \centering
        \includegraphics[width=\linewidth]{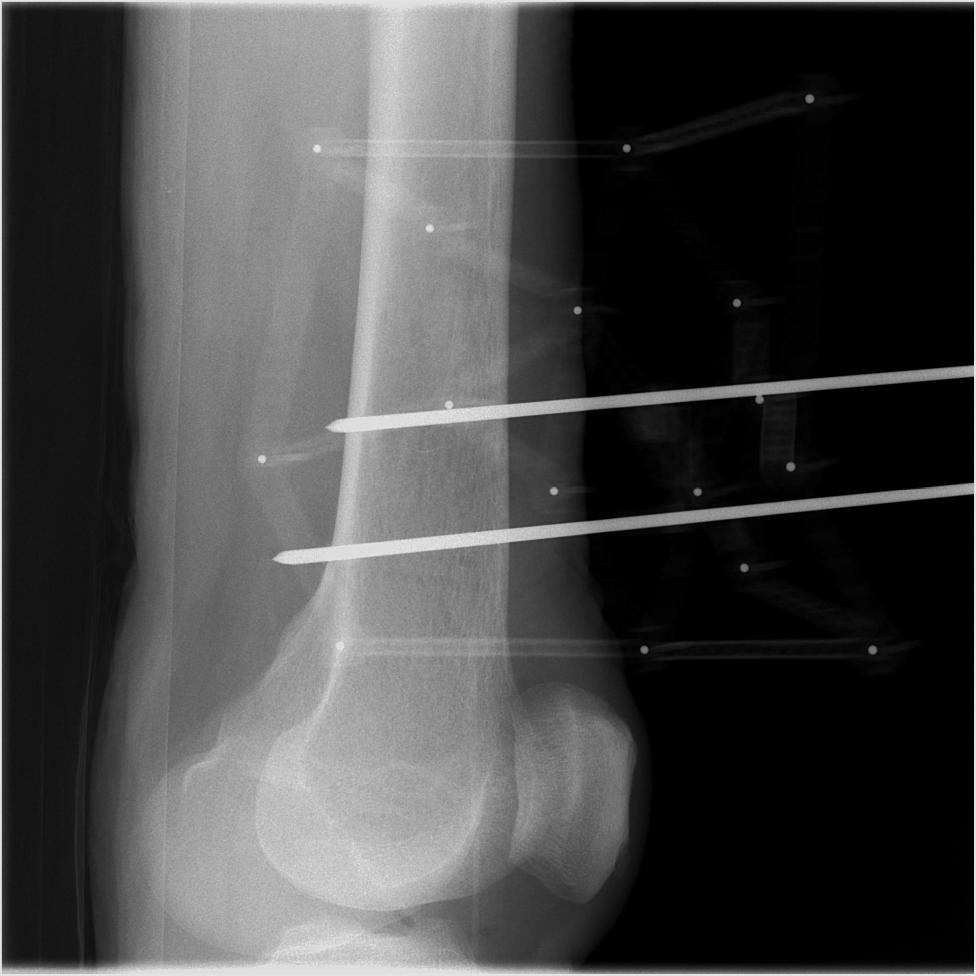}
    \end{subfigure}
    \hspace{-2mm} 
    \begin{subfigure}{0.225\linewidth}
       \centering
        \includegraphics[width=\linewidth]{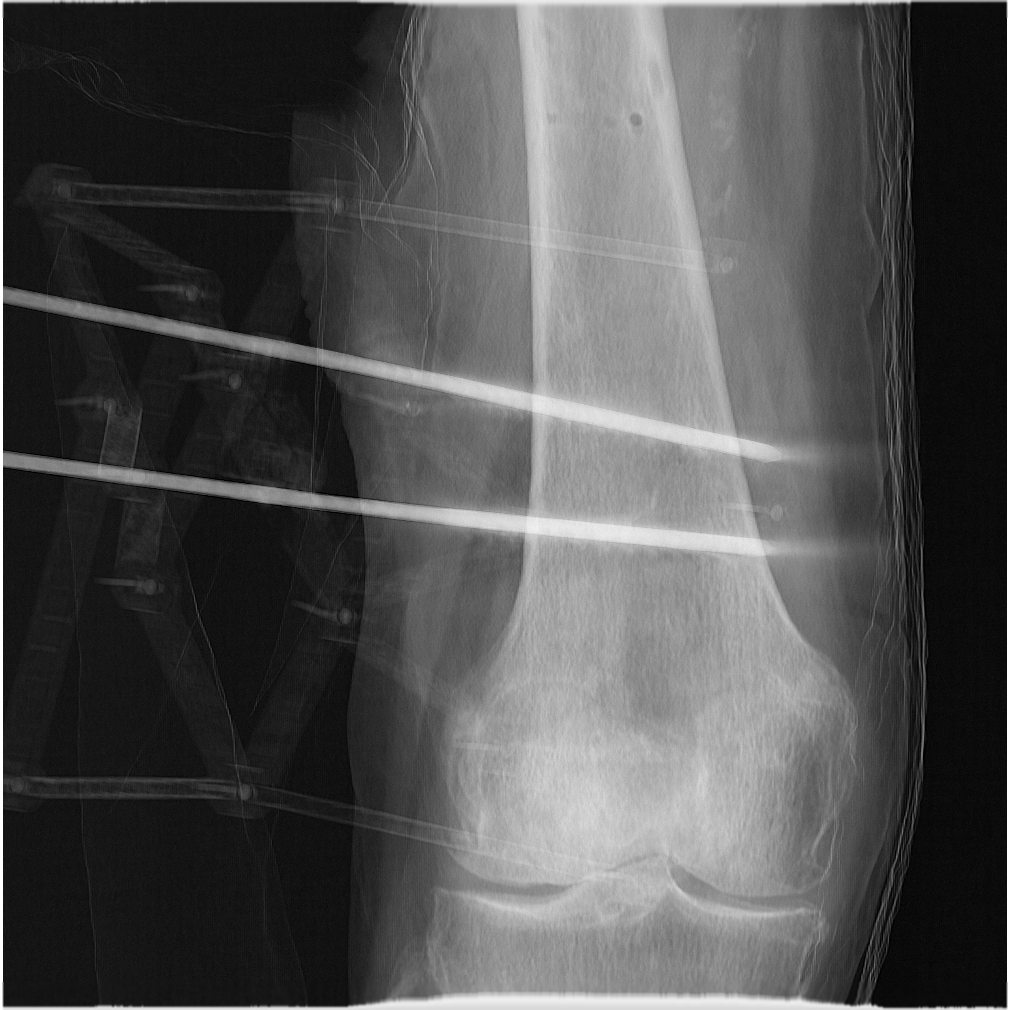}
    \end{subfigure}
    \hspace{2mm} 
    \begin{subfigure}{0.225\linewidth}
        \centering
        \includegraphics[width=\linewidth]{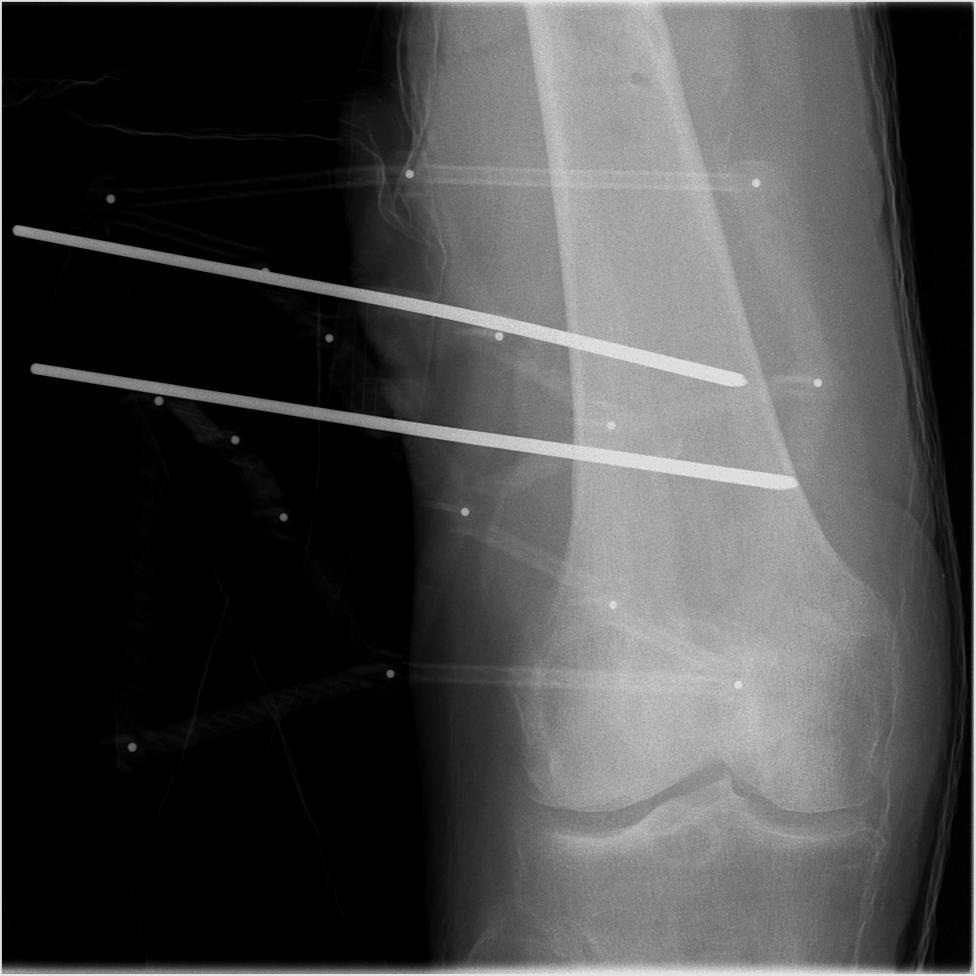}
    \end{subfigure}
    \hspace{-2mm} 
    \begin{subfigure}{0.225\linewidth}
        \centering
        \includegraphics[width=\linewidth]{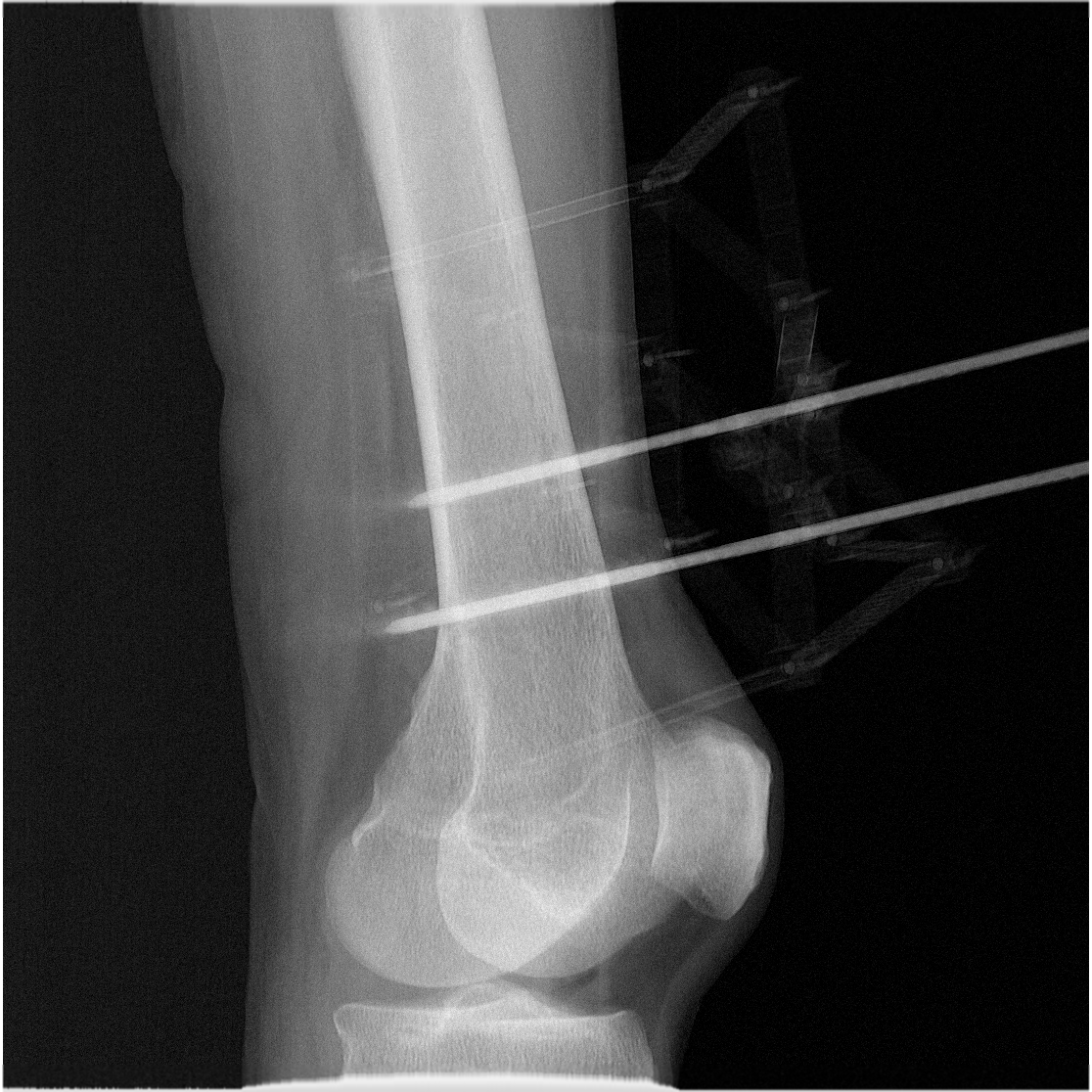}
    \end{subfigure}
    \caption{Two examples of DRR augmentation. Left: Augmented DRR with added noise, adapted contrast, matched artifacts (white edge regions). Right: Real X-ray image of a similar pose}
    \label{fig:training_data}
\end{figure}

{A critical step in improving the contour model predictions involved augmenting the generated DRRs at various stages to closely resemble real X-ray images. The DRRs were generated using 3D Slicer \cite{slicer3d, FEDOROV20121323}, a robust and open-source toolbox for medical imaging. Unlike real X-ray images, which exhibit significant noise and visual artifacts (see Figure \ref{fig:training_data}), the generated DRRs are typically very clean and artifact-free (see Figure \ref{fig:training_data_synth}).
To bridge this gap, our data augmentation strategy focused on introducing various levels of Gaussian noise, random contrast adjustments, applying small random rotations, and simulating white edge artifacts.} 
{These standard techniques have been shown to effectively bridge the imaging domain gap between synthetic and real-world data, as supported by prior work \cite{cossio2023augmentingmedicalimagingcomprehensive}. Despite these domain gaps, our study demonstrates that our model, trained exclusively on synthetic data, generalizes exceptionally well to real X-ray images. This claim is supported by Figures 3, 4, and 5 of the main article which show results obtained on real X-ray images.}

{
\section{Memory Usage}
\label{sec:memory}
The memory footprint of our method consists of two main components: the registration optimization and the nnU-Net inference. During the registration process, the memory usage is minimal, requiring approximately 2.8GB of RAM. The nnU-Net inference, which is dynamically managed, utilizes around 18GB of GPU memory on an NVIDIA Quadro P6000 with 24GB of dedicated memory. These measurements reflect current implementation settings without specific memory optimization. Future work may explore techniques to further reduce the memory requirements.} 
\end{appendices}

\bibliography{sn-bibliography}

\end{document}